\PassOptionsToPackage{x11names,dvipsnames}{xcolor}
\documentclass[11pt]{article}

\usepackage[preprint]{acl}

\usepackage{times}
\usepackage{latexsym}
\usepackage{xcolor}
\usepackage[T1]{fontenc}
\usepackage[utf8]{inputenc}

\usepackage{microtype}

\usepackage{inconsolata}

\usepackage{inconsolata}
\usepackage{amsmath}
\usepackage{cleveref}
\usepackage{graphicx}
\definecolor{grey}{rgb}{0.5,0.5,0.5}
\usepackage{tabularx}
\usepackage{siunitx}
\usepackage{multirow}
\usepackage{amsfonts}
\usepackage{amssymb}
\usepackage{xspace}
\usepackage{lipsum}
\usepackage{adjustbox}
\usepackage{booktabs}
\usepackage{makecell}
\usepackage{url}
\usepackage{enumitem}
\setlist[enumerate]{topsep=0pt, partopsep=0pt, parsep=0pt, itemsep=0pt}
\setlist[itemize]{topsep=0pt, partopsep=0pt, parsep=0pt, itemsep=0pt}
\usepackage{listings}
\usepackage{xcolor}
\usepackage{colortbl}
\usepackage[most]{tcolorbox}
\usepackage{caption}
\usepackage{float}
\usepackage{placeins}
\usepackage{tikz}
\usepackage{pgfplots}
\usepackage{subcaption}
\pgfplotsset{compat=1.18}
\usepgfplotslibrary{groupplots}
\usepackage{amsthm}
\usepackage{alltt}
\newtheorem{definition}{Definition}
\usepackage{diagbox}

\newcommand{\avg}{\rowcolor{black!8}\textit{Average}}

\newcommand*\samethanks[1][\value{footnote}]{\footnotemark[#1]}

\definecolor{plannerblue}{HTML}{1F3B73} %1F3B73
\definecolor{pddlorange}{HTML}{E69F00} %E69F00
\definecolor{cpsatgreen}{HTML}{009E73} %009E73
\definecolor{repairpurple}{HTML}{CC79A7}

\newcommand{\planner}{\textcolor{plannerblue}{\textsc{Planner}}\xspace}
\newcommand{\pddl}{\textcolor{pddlorange}{\textsc{PDDL2.1 Formalizer}}\xspace}
\newcommand{\cpsat}{\textcolor{cpsatgreen}{\textsc{CP-SAT Formalizer}}\xspace}
\newcommand{\staterepair}{\textcolor{repairpurple}{\textsc{State-aware Repair}}\xspace}

\newcommand{\Formalizer}{Formalizer\xspace}
\newcommand{\Formalizers}{Formalizers\xspace}

\newcommand{\drexel}{%
  \hspace{1pt}
  \begingroup\normalfont
  \includegraphics[height=1.3\fontcharht\font`\B]{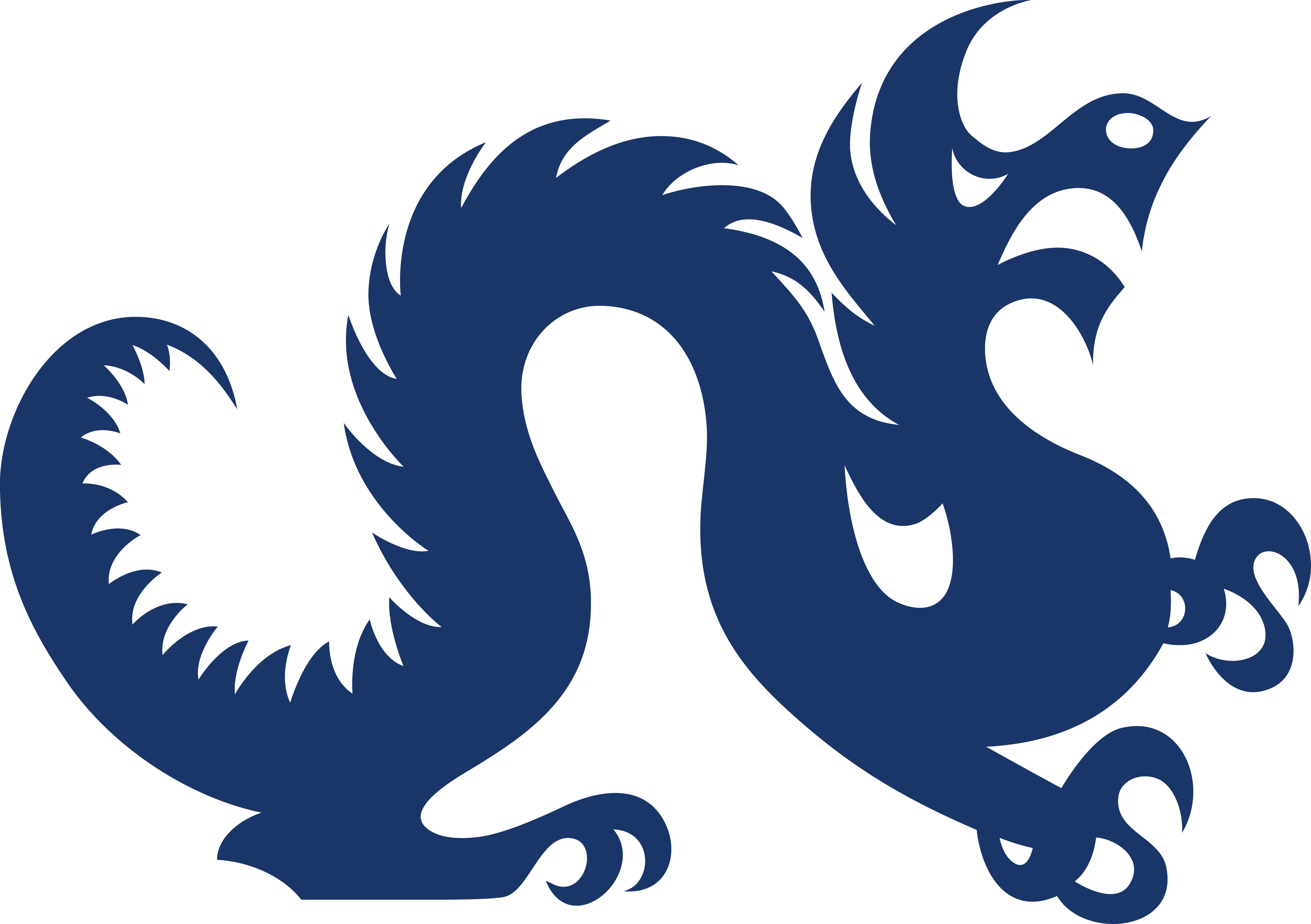}%
  \endgroup
  \hspace{1pt}
}
\newcommand{\asu}{%
  \hspace{1pt}
  \begingroup\normalfont
  \includegraphics[height=1.3\fontcharht\font`\B]{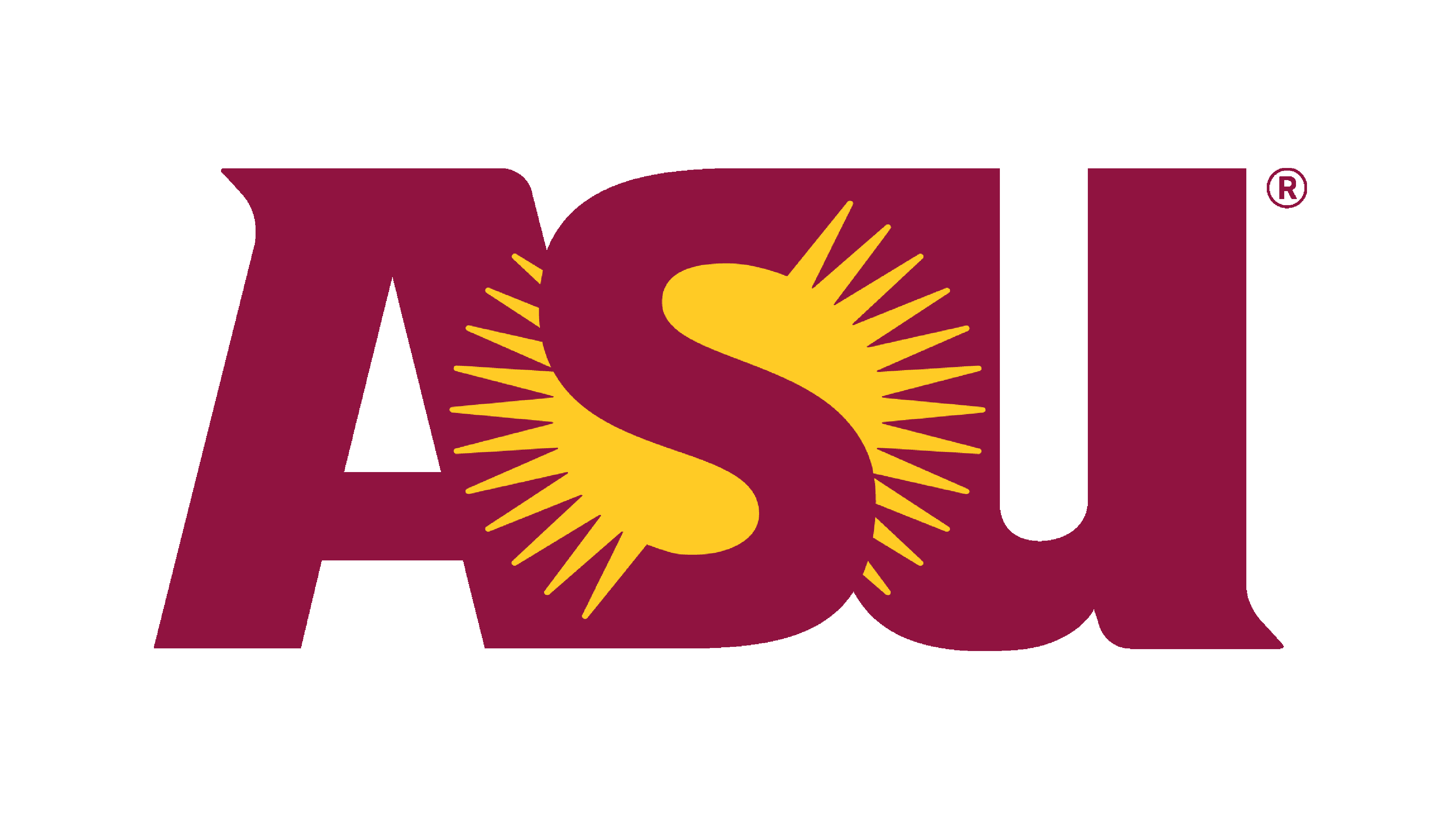}%
  \endgroup
  \hspace{1pt}
}

\lstdefinestyle{mystyle}{
    basicstyle=\ttfamily\small,
    breaklines=true,
    frame=single,
    backgroundcolor=\color{gray!5},
    keywordstyle=\color{black},
    commentstyle=\color{gray},
    stringstyle=\color{black},
    showstringspaces=false,
    tabsize=2,
    rulecolor=\color{black},
    framesep=6pt, 
    columns=fullflexible,
}
\title{Robust Asynchronous Planning via Auto-Formalization}

\author{Jiayi Zhang\drexel\thanks{\hspace{4pt} Equal contribution; work done as an intern at Drexel University.} \enspace Jianing Yin\drexel\samethanks \enspace Ben Zhou\asu \enspace Li Zhang\drexel \\
  \drexel Drexel University \hspace{4pt} \asu Arizona State University \\
  {\tt jiayizx@uchicago.edu|Harry.Zhang@drexel.edu}
}

\begin{document}
\maketitle
\begin{abstract}
LLMs can plan by generating schedules directly as a \emph{Planner} or by translating a task into a formal specification for a solver as a \emph{Formalizer}. Most real-world tasks are \emph{asynchronous}, with non-uniform durations, concurrency, and execution-time changes, yet existing benchmarks hardly cover them. We unify asynchronous planning under one formulation and release three benchmarks spanning task-graph scale, grounded state and resource constraints, and execution-time events.
% \footnote{We include the code and benchmark data in both the Software and Data fields of our ARR submission.}
Surprisingly, PDDL2.1, the formalism built for temporal planning, is the weakest pipeline. On AsyncPlan-XXL it averages 30.8\% plan accuracy, against 81.7\% for the Planner and 94.0\% for a CP-SAT \Formalizer. The gap widens with scale, and at 100 actions the three pipelines reach 9.0\%, 44.5\%, and 82.5\%, respectively. The formalism is not to blame. Compiled from the same model-extracted intermediate representation, PDDL2.1 and CP-SAT reach nearly identical accuracy (81.6\% vs.\ 81.5\%), and gold task graphs reach 100\% for both. 
What remains is a bottleneck shared by every pipeline, recovering the task graph from language.
Execution-time events further degrade every pipeline, yet updating only event-induced constraints reaches 81.6\% versus 41.6\% for full regeneration.
Robust asynchronous planning thus hinges not on the expressiveness of the target formalism but on how reliably a model can write and edit it, and a well-chosen interface can let a weaker model match a stronger one.
\footnote{Code and data are available at \url{https://github.com/jiayizx/async_planning}.}
\end{abstract}

\section{Introduction}
\label{sec:introduction}
\begin{figure*}[t!]
    \centering
    \includegraphics[width=\textwidth]{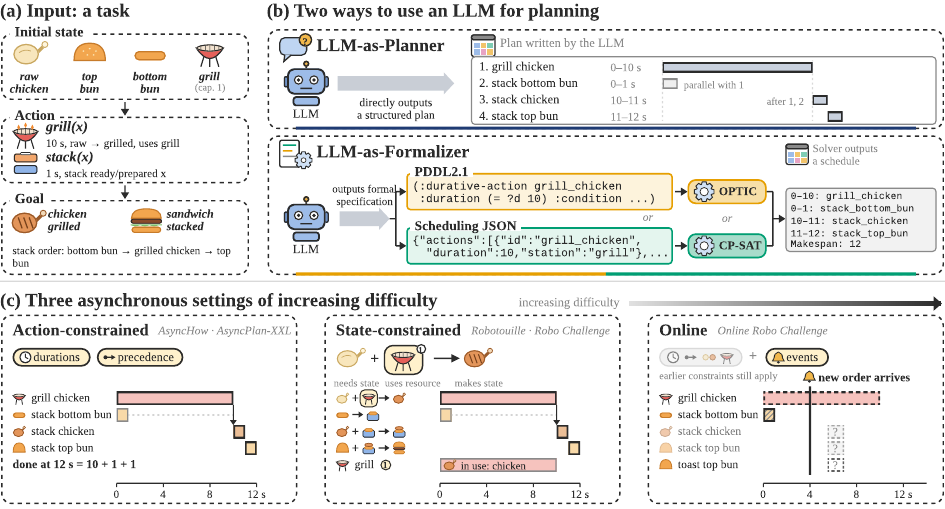}
    \caption{\textbf{LLM-as-Planner vs. LLM-as-Formalizer} for asynchronous planning. \textbf{(a)} Given a natural-language task with initial state, durative actions, and goal. \textbf{(b)} The \emph{Planner} directly writes a structured plan, while the \emph{Formalizer} writes a formal specification that a solver turns into a schedule. \textbf{(c)} We evaluate both across three settings of increasing difficulty, testing whether the model can \emph{recover} durations and precedence (action-constrained), \emph{ground} resource and state changes (state-constrained), and \emph{maintain} a valid plan as events change mid-execution (online).}
    \label{fig:teaser}
\end{figure*}

Large Language Models (LLMs) have been studied extensively for planning, both as end-to-end \textbf{Planners} that generate action sequences directly from a natural language task~\cite{valmeekam2024planbench,kambhampati2024position} and as \textbf{\Formalizers} that translate the task into a formal specification for an external solver~\cite{liu2023llmp,hao2025planning,huang2025languagemodelplannerformalizer} (Figure~\ref{fig:teaser}(b)).
Yet this attention has focused mostly on \emph{synchronous} planning, where actions are atomic and execute one at a time.
Real-world tasks such as cooking a multi-course meal~\cite{gonzalez-pumariega2025robotouille}, operating a household robot~\cite{puig2018virtualhome} or coordinating train dispatch across a rail network~\cite{mohanty2020flatlandrlmultiagentreinforcement} are instead \emph{asynchronous}: actions take non-uniform durations, run in parallel, interact through shared state and resource constraints, and the task can change mid-execution.
Recent benchmarks show that current LLMs struggle here: Planners must reason about parallel, interacting actions rather than a single sequence, while \Formalizers must preserve durations, dependencies, and states faithfully in their formal specification~\cite{10.5555/3692070.3693283,gonzalez-pumariega2025robotouille,zhang2026paracooktimeefficientplanningmultiagent}.
% \harry{Say why planner (reasoning with constraints) and formalizer (language support and code generation challenge) each fail, even when they're good with synchronous planning.}

Existing asynchronous benchmarks each cover only a slice of this problem space. 
AsyncHow~\cite{10.5555/3692070.3693283} studies action duration and dependency on short wikiHow tasks~\cite{koupaee2018wikihow}; Robotouille~\cite{gonzalez-pumariega2025robotouille} has ten generic tasks that do not separate the challenges we target; ParaCook~\cite{zhang2026paracooktimeefficientplanningmultiagent}
studies the multi-agent setting, but assumes fixed task descriptions.
% \harry{what's the downside of ParaCook compared to ours?}. 
None jointly evaluates problem scale, grounded states, resource constraints, and dynamically changing goals.
Richer formalisms can encode durations, concurrency, resources and online updates, but they also introduce syntactic and semantic errors~\cite{huang2025languagemodelplannerformalizer}, so we study multiple target languages rather than assuming one suffices.
% Moreover, the natural fit for asynchronous planning, an LLM as Formalizer rests on a critical assumption: the generated formal artifact must \emph{faithfully} encode the task. \harry{I suggest not focus on the faithfulness issue as we don't explicitly address this either; see my proposal in the next comment.} A syntactically valid PDDL file may still omit dependencies, misstate durations, or drop goals, and prior work on classical planning shows that the gap between solvable and correct is large~\cite{huang2025languagemodelplannerformalizer}. \harry{Should clearly say that previous work in formalizers only tried to generate PDDL1.2 which does not support async planning, while Cassie's ACL2026 paper shows that generating PDDL3.1 causes both syntactic and semantic errors. This motivates our attempt in multiple suitable languages which is non-trivial.}

As shown in Figure~\ref{fig:teaser}(c), we unify asynchronous planning into three settings of increasing difficulty: \emph{action-constrained} (durations and precedence), \emph{state-constrained} (grounded states and resource constraints), and \emph{online} (execution-time events).
We introduce one benchmark per setting. \textbf{AsyncPlan-XXL} scales AsyncHow-style tasks up to 100 actions, \textbf{Robo Challenge} builds on Robotouille with denser state and resource constraints, and \textbf{Online Robo Challenge} adds events that revise the task mid-execution.
Across four LLMs, we compare the Planner with a \Formalizer in two target languages, PDDL2.1 solved by the OPTIC temporal planner~\cite{benton2012temporal} and a more general CP-SAT scheduling specification~\cite{perron2023cp}, and we score plan accuracy together with five \emph{faithfulness diagnostics} that locate formalization failures.
% (Figure~\ref{fig:method-dataset-overview}) 
% For each setting, we introduce a new benchmark based on an existing benchmark that isolates the corresponding challenge.
% \harry{Since multi-agent is not discussed above in ``grounded planning'', I suggest not introducing it now.}
% \textbf{Online Robo Challenge} further adds execution-time events that revise the task specification during planning. 
%This comparison allows us to test both whether formalization helps and which representation is reliable for asynchronous planning.
% , including dependency precision and recall, goal completeness, action-count match, and duration match against the gold action graph.
% \harry{I suggest consistenly using the terms Formalizer and Planner throughout the paper; succinct and clear.}
% which include a PDDL2.1 Formalizer~\cite{benton2012temporal} paired with the OPTIC temporal planner and a CP-SAT Formalizer~\cite{perron2023cp} paired with a constraint programming solver. \ben{I suggest we further clarify the differences and motivate why we need these two. This will help make sense to why CP-SAT is so much better than PDDL.}
% Beyond end-task plan accuracy, we also evaluate the generated formal representations of the tasks with \emph{faithfulness diagnostics}, which include dependency precision and recall, goal completeness, action-count match, and duration match against the gold action graph. \ben{I am not sure if the intro has defined or is consistent with the ``unifying action-, predicate-, and Online settings'' aspect as in the abstract?}

Our findings are as follows. 
First, \Formalizer is substantially more robust than Planner when the formal specification matches the task: as task graphs grow, the CP-SAT \Formalizer barely degrades, whereas the Planner holds up for some models and collapses for others (\S\ref{sec:rq1}).
Second, PDDL2.1, though designed for exactly this kind of temporal planning, is the weakest pipeline at scale, trailing CP-SAT by more than 60 points on average. We trace the gap not to the formalism but to the act of writing it (\S\ref{sec:rq2}). Compiling the gold task graph reaches 100\% accuracy in both PDDL2.1 and CP-SAT, and compiling a model-extracted graph reaches 81.6\% and 81.5\%, which leaves recovering the graph from language as the one bottleneck every pipeline shares.
Third, when the task changes mid-execution, updating only the affected constraints roughly doubles plan accuracy compared with rewriting the specification from scratch (\S\ref{sec:rq3}).
Together, these results show that, for the pipelines and benchmarks studied here, robust asynchronous planning depends on how well the formal specification matches the task structure, not on how much the target language can express.

\section{Related Work}
\label{sec:related_work}

% \ben{we can add some works on classic temporal planning, such as Temporal constraint networks,  PSPLIB and Resource-Constrained Project Scheduling.}

\paragraph{Planning with LLMs.}
Whether LLMs can plan robustly remains debated. PlanBench~\citep{valmeekam2024planbench} and follow-up evaluations~\citep{kambhampati2024position,valmeekam2024llmscantplanlrms,huang-zhang-2025-limit} show that even strong models struggle on classical International Planning Competition domains, and some apparent successes may reflect shortcuts or memorization rather than genuine planning ability~\citep{stechly2024chain,shojaee2026illusion}. 
Beyond single-shot planning, other work interleaves LLM reasoning with environment feedback~\citep{yao2022react,guan2023leveraging}, and recent surveys map this expanding space~\citep{wei-etal-2025-plangenllms,tantakoun-etal-2025-llms}. Across this landscape, however, evaluation has concentrated on \emph{synchronous} planning, where actions are atomic and execute one at a time.
The \emph{asynchronous} setting, with non-uniform durations and concurrency under temporal, state and resource constraints, has received much less attention, despite being the one long studied by temporal constraint networks~\citep{dechter1991temporal} and the PSPLIB/RCPSP literature on resource-constrained scheduling~\citep{kolisch1997psplib,hartmann2010survey}.
Despite this history, asynchronous planning has only recently entered LLM evaluations.
AsyncHow~\citep{10.5555/3692070.3693283} studies duration-and-precedence problems derived from wikiHow, Robotouille~\citep{gonzalez-pumariega2025robotouille} extends this to a grounded kitchen environment, and ParaCook~\citep{zhang2026paracooktimeefficientplanningmultiagent} investigates a multi-agent cooking variant.
Together, these benchmarks point to the importance of asynchronous planning, but they cover only parts of the problem space: none jointly tests scale, grounded state and limited shared resources, and online adaptation to events such as new tasks or resource failures.
This gap motivates our work.

\paragraph{Formalizing with LLMs.}
Rather than planning directly, a complementary line of work uses LLMs as \emph{\Formalizers} that translate natural-language tasks into formal specifications for an external solver.
This idea spans many domains: program-aided language models translate word problems into executable code~\citep{gao2023pal, vashishtha2025executable}, mathematical statements are formalized in Lean~\citep{ying2024lean, hubert2025olympiad}, and natural-language queries are translated into domain-specific languages such as SQL or LTL~\citep{xie2023translating,yang2024plug}.
For classical planning specifically, Formalizer has been instantiated with PDDL~\citep{liu2023llmp,silver2024generalized,wong2023learning,zuo-etal-2025-planetarium,kagitha2025addressingchallengesplanninglanguage,hu-etal-2025-text2world, sun2023adaplanner}, with constraints encoded in PDDL3, LTL, or SMT~\citep{huang2025languagemodelplannerformalizer,guo2025castl}, and with constraint-satisfaction or general-purpose programming as the target language~\citep{amonkar2025naturallanguageplanningcoding,hao2025planning}.
However, existing \Formalizer work almost exclusively targets synchronous planning.
Whether LLMs can faithfully translate asynchronous tasks into temporal formalisms such as PDDL2.1 or CP-SAT, and whether the choice of representation interacts with problem structure, remains largely unexplored.
We close this gap by evaluating Planner and \Formalizer under a single asynchronous planning formulation across three settings of increasing difficulty, showing that how well the formal specification matches the task structure, more than how much the target language can express, drives both robustness and scalability.

\section{Task Formulation}
\label{sec:task_formulation}

We characterize the space of asynchronous planning problems that share four properties:
(i) actions have non-uniform, non-zero durations;
(ii) actions may execute in parallel with preconditions and effects;
(iii) constraints restrict actions or states based on \citet{huang2025languagemodelplannerformalizer} which in turn follows the convention of PDDL3 \cite{gerevini2005pddl3};
(iv) a plan is a schedule of actions that achieve the goal state, minimize the makespan (total time), and respect all constraints.
Continuous \cite{fox2006pddl} and stochastic effects \cite{younes2004ppddl} are out of scope.

\paragraph{Problem Definition.}
We define an asynchronous planning instance as
$\mathcal{P} = \langle s_0, g, \mathcal{A}, \mathcal{T}, \mathcal{C} \rangle$,
where $s_0 \in \mathcal{S}$ is the initial state. $g: \mathcal{S}\rightarrow \mathbb{B}$ is a goal condition. $\mathcal{A}=\{a_1,\ldots,a_n\}$ is the set of actions and $\mathcal{T}: \mathcal{A}\rightarrow \mathbb{R}^{+}$ maps each action to its duration. 
At the end of the execution, an action modifies the state $s_{i+1}=a_j(s_i)$.
$\mathcal{C}: (\mathcal{S}, \mathcal{A}\rightarrow \mathbb{B})$ is a set of constraints that dictate whether an action can be executed given a state, such as a pre-condition. 
% \jiayi{H -> time limit (more specific definition); C is a boolean a function of s and a, given current s_i whether s_{i+1} can be executed; H -> time limit}

A plan is a schedule
$\pi = \{(a_i, t_i)\}_{i=1}^{N}$,
where $a_i$ is an action and $t_i \in \mathbb{R}^{\geq 0}$ is the start time.
The plan is valid if it achieves the goal $g$ and respects every constraint in $\mathcal{C}$.
The default objective minimizes the makespan $T_{\mathrm{total}}(\pi)=\max_i \left(t_i+\mathcal{T}(a_i)\right)$; one family of instances instead maximizes completed-goal reward under a deadline and resource limits.

\paragraph{Constraint decomposition.}
We partition the constraint set $\mathcal{C}$ of an asynchronous
planning instance into three disjoint classes,
\[
\mathcal{C}\ =\ \mathcal{C}_A\ \cup\ \mathcal{C}_S\ \cup\ \mathcal{C}_R,
\]
where
$\mathcal{C}_A \subseteq \mathcal{A} \times \mathcal{A}$ is the
set of \emph{action} (precedence) constraints,
$\mathcal{C}_S$ is the set of \emph{state} predicates
(action preconditions $\psi_a^{\mathrm{pre}}$, effects
$\psi_a^{\mathrm{eff}}$, and goal predicates over $\mathcal{S}$),
and $\mathcal{C}_R$ is the set of \emph{resource} predicates
(the set of shared executors $\mathcal{K}$ with their capacities, and shared devices
$\{r : C_r \in \mathbb{Z}^{+}\}$). We write
$\mathcal{C}_P := \mathcal{C}_S \cup \mathcal{C}_R$ for the
predicate-level component, and we call $\langle \mathcal{A}, \mathcal{T}, \mathcal{C}_A \rangle$ the \emph{task graph} of an instance: which actions it contains, how long each one takes, and which must finish before which.

\paragraph{Problem Categories.}
Each setting is defined by which classes of $\mathcal{C}$ are active and whether the instance is fixed or revised during execution.
The settings form a strict inclusion $\mathcal{P}_A \subset \mathcal{P}_S \subset \mathcal{P}_O$: $\mathcal{P}_A$ activates only action constraints, $\mathcal{P}_S$ additionally activates predicate constraints, and $\mathcal{P}_O$ generalizes $\mathcal{P}_S$ to evolving specifications. 
\S\ref{sec:experimental_setup} pairs $\mathcal{P}_A$ with AsyncHow and AsyncPlan-XXL, $\mathcal{P}_S$ with Robo Challenge and $\mathcal{P}_O$ with Online Robo Challenge.

% as one of three constraint settings, defined by which subsets of $\mathcal{C}$ are active and whether the instance is fixed or evolving over execution time. 
% The settings form a strict inclusion
% $\mathcal{P}_A \subset \mathcal{P}_S \subset \mathcal{P}_O$:
% $\mathcal{P}_A$ activates only action constraints, $\mathcal{P}_S$
% additionally activates predicate constraints, and $\mathcal{P}_O$
% generalizes $\mathcal{P}_S$ to evolving specifications.
% \S\ref{sec:experimental_setup} pairs each setting with a benchmark:
% $\mathcal{P}_A$ with AsyncHow and AsyncPlan-XXL, $\mathcal{P}_S$
% with Robo Challenge, and $\mathcal{P}_O$ with Online Robo Challenge.

\begin{definition}[Action-Constrained]
$\mathcal{P}_A$ is a special case of $\mathcal{P}$ with
$\mathcal{C}\subseteq \mathcal{A} \times \mathcal{A}$,
$s_0 = \emptyset$, and
$g = \bigwedge_{a \in \mathcal{A}} \mathrm{done}(a)$.
A plan reduces to $\pi = \{(a_i, t_i)\}_{i=1}^{N}$ and is valid iff
(i) every $a \in \mathcal{A}$ appears in $\pi$;
(ii) $\forall (a_p, a_q)\in \mathcal{C}_A:\
t_p + \mathcal{T}(a_p) \le t_q$.
The optimum $T_{\mathrm{total}}^{\star}$ is the critical-path makespan
of the DAG induced by $\mathcal{C}_A$.
\end{definition}
Intuitively, action-constrained planning means that each action executes exactly once, with precedence relations specifying the order of execution: the only pre-condition of an action is the completion of one or more other actions.

\begin{definition}[State-Constrained]
$\mathcal{P}_S$ is a general case with
$\mathcal{C}\subseteq \mathcal{S} \times \mathcal{A}$, and concrete $s_0, g$. Let
% $s : [0,H]\to\mathcal{S}$ be the state trace induced by applying
$s(t)$ be the state trace induced by applying
every action $a$ at time $t_a$. A plan
% $\pi = \{(a_i, k_i, t_i)\}_{i=1}^{N}$ is valid iff
% (i) the action pre-conditions of $\mathcal{P}_S$ hold;
% (ii) $\forall a \in \pi:\ a(s(t_a^{-}))=T$ and
% $s(\max_i t_i + \mathcal{T}(a_i)) \vDash g$.
$\pi = \{(a_i, k_i, t_i)\}_{i=1}^{N}$, where $k_i \in \mathcal{K}$ is the executor assigned to $a_i$, is valid iff
(i) every action's pre-conditions hold in the state at its start time, $a_i(s(t_i^{-}))$; and
(ii) the final state satisfies the goal, $s(\max_i t_i + \mathcal{T}(a_i)) \vDash g$.
\end{definition}

Intuitively, state-constrained planning means that actions modify a grounded world state and must satisfy concrete pre-conditions and goals: pre-conditions, effects, and the goal are all sets of grounded predicates, as in classical planning.

\begin{definition}[Online]
% $\mathcal{P}_O$ is an extension of $\mathcal{P}_S$, which is a pair of $\langle \mathcal{P}_S^{(0)},\ \{e_t\}_{t=1}^{T}\rangle$,
$\mathcal{P}_O$ is an extension of $\mathcal{P}_S$, which is a pair of $\langle \mathcal{P}_S^{(0)},\ \{e_t\}_{t=1}^{M}\rangle$,
where $\mathcal{P}_S^{(0)}$ is an initial state-constrained
instance and each event is a revision
$e_t : \mathcal{P}_S^{(t-1)} \to \mathcal{P}_S^{(t)}$
$1\le t \le M$) arriving at execution time $\tau_t$. The agent commits to a sequence of partial
schedules $\pi^{(0)}, \pi^{(1)}, \dots, \pi^{(M)}$ such that, for
every $t$, $\pi^{(t)}$ is valid for $\mathcal{P}_S^{(t)}$ given the
prefix of actions executed before $e_t$. For ordinary tasks, the objective is the
makespan of the final schedule $\pi^{\star} = \pi^{(M)}$ measured from $t=0$.
Reward variants instead maximize completed-goal reward under a deadline and
inventory limits; makespan is a secondary efficiency measure.
\end{definition}
Intuitively, the requirement may dynamically change \textit{during} the execution of a plan, where the change $e_t$ is described in natural language. 

%\jiayi{check which element changes during the execution in our benchmark; add H}

\section{Experimental Setup}
\label{sec:experimental_setup}

\begin{figure}[t]
    \centering
    \includegraphics[width=\columnwidth]{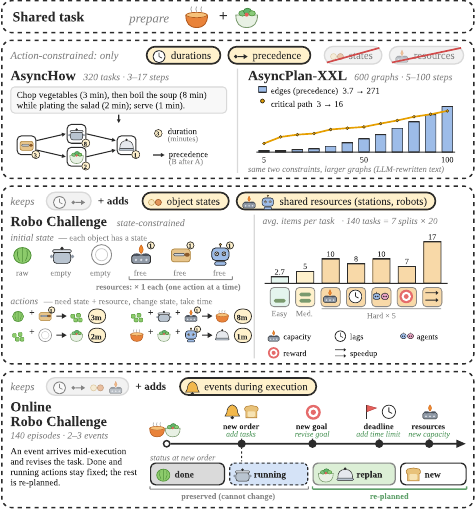}
    \caption{Three asynchronous planning settings covered by our benchmarks. AsyncHow and AsyncPlan-XXL test action-level temporal reasoning (durations and precedence). Robo Challenge adds executable grounding: object states, action schemas, resources, and agents. Online Robo Challenge extends this with execution-time events, requiring the pipeline to preserve completed and running actions while repairing the remainder.
    \vspace{-1em}
    }
    \label{fig:method-dataset-overview}
\end{figure}

We evaluate LLM-based asynchronous planning under the three settings of \S\ref{sec:task_formulation}, each paired with a benchmark that stresses it: recovering durations and precedence relations, planning under state and resource constraints, and repairing plans after execution-time changes. Figure~\ref{fig:method-dataset-overview} shows how the benchmarks instantiate these settings.

\paragraph{Benchmarks.}
\label{sec:benchmarks}
For the action-constrained setting, AsyncHow~\cite{10.5555/3692070.3693283} contains 320 small, shallow duration-and-precedence tasks derived from wikiHow~\cite{koupaee2018wikihow}. We introduce \textbf{AsyncPlan-XXL}, with 600 synthetic dependency graphs of 5 to 100 actions (50 per size), sampled by rank-based edge sampling~\cite{wang2024languagemodelssolvegraph,zhang2025planovergraphparallelablellmagent} that rejects insufficient connectivity or trivial critical paths and enforces a minimum width for parallelism. Gemini-3-Flash then rewrites each graph into a natural-language task with concrete step descriptions and realistic durations.
For the state-constrained setting, Robotouille~\cite{gonzalez-pumariega2025robotouille} offers 10 grounded kitchen tasks, but all three pipelines score near zero (Table~\ref{tab:async_robo_full_results}), largely due to low-level navigation and state bookkeeping (Appendix~\ref{app:robotouille}). We thus introduce \textbf{Robo Challenge}, which retains grounded constraints and executable action schemas while abstracting these away. Its 140 tasks form seven 20-task splits: Easy, Medium, and five Hard splits each stressing one constraint type: shared station capacity, temporal lags, conflict-free multi-agent assignment, reward optimization under deadlines and inventory limits, and high parallel speedup.
\textbf{Online Robo Challenge} augments each Robo Challenge instance with an event stream $\{e_t\}_{t=1}^{M}$ that revises the instance during execution (e.g., adding tasks, revising goals, introducing new constraints), requiring pipelines to preserve completed and running actions when replanning. Prior work studies asynchronous planning either with fixed task descriptions~\citep{10.5555/3692070.3693283,gonzalez-pumariega2025robotouille,zhang2026paracooktimeefficientplanningmultiagent} or replanning under state changes without duration semantics~\citep{valmeekam2024planbench}; to our knowledge, Online Robo Challenge is the first to combine both. Full statistics for all three benchmarks appear in Appendix~\ref{app:dataset_stats}.

\paragraph{Pipelines.}
As shown in Figure~\ref{fig:teaser}, we compare three pipelines, all reading the same natural-language task but differing in what the LLM outputs and what turns that output into a schedule. 
\planner maps the task directly to a structured plan with no solver involved; a deterministic program computes the makespan from the predicted plan. \pddl instead generates PDDL2.1 for OPTIC~\citep{benton2012temporal}, which expresses durative actions, predicates, and goals naturally but requires more elaborate encoding for resource constraints and state management. 
\cpsat generates scheduling JSON with explicit precedence relations and resource capacities for CP-SAT~\citep{perron2023cp}. 
Both \Formalizers may rewrite their formalization up to three times, since syntax checking is cheaper than full planning.
% On AsyncHow and AsyncPlan-XXL, both \Formalizers may rewrite their specification up to three times when it fails to parse or validate, when the solver finds no plan, or when the solved makespan is wrong; the \planner makes a single attempt. On Robo Challenge, the CP-SAT \Formalizer makes a single attempt and the PDDL2.1 \Formalizer gets one repair round from the simulator's error message.

\paragraph{Evaluation.}
Our primary metric, \textbf{plan accuracy}, counts an instance as correct iff $\pi \models C \wedge g$ and $T_{\text{total}}(\pi) = T^\star$, where $\pi$ is the predicted schedule and $T^\star$ is the gold optimal makespan. On AsyncPlan-XXL it requires every action to appear once with its stated duration, every gold precedence constraint to hold, and the recomputed makespan to be optimal. Robo Challenge additionally checks grounded execution, temporal lags, and goals, and requires optimal reward on Hard Optimization. Online, a plan must also complete the event protocol at the true arrival times while preserving ongoing actions. AsyncHow's annotations only support the numerical objective, so we report \textbf{makespan accuracy} there, i.e.\ whether $T_{\text{total}}(\pi) = T^\star$, which is not comparable to plan accuracy.

Plan accuracy scores only the final schedule. To locate errors before the solver runs, we introduce five \textbf{faithfulness diagnostics} that score what the model itself wrote.
They compare the task graph and goals in its output with the gold ones on AsyncPlan-XXL, covering the action count, per-action durations, goal-action coverage, and the precision and recall of precedence edges. 
We parse each output and match its actions to the gold actions, and a diagnostic scores zero if either step fails. We call an output \textbf{verified} when all five diagnostics give full marks. We use the diagnostics in \S\ref{sec:rq2} and report them in full in Appendix~\ref{app:error_analysis}.
% Beyond plan accuracy, five \textbf{faithfulness diagnostics} on AsyncPlan-XXL compare the task structure an output writes down, before any solver runs, with the gold actions $A$, durations $T$, constraints $C_A$ and goals $g$: action count, per-action durations, goal-action coverage, and precedence precision and recall. A diagnostic that cannot be computed, because the output cannot be parsed or its actions cannot be matched to gold actions, scores zero. \S\ref{sec:rq2} uses them, and Appendix~\ref{app:error_analysis} reports them in full.
% \jiayi{check the terminology alignment; what 34.3 mean here?}
% \cathy{Fixed the sentence and aligned names with Table~\ref{tab:faithfulness}. 34.3 is the share of PDDL2.1 records on which all five diagnostics can be computed. It is low because 34.7\% of outputs fail to parse (mostly missing \texttt{:parameters}) and 26.9\% use a shared \texttt{?s} parameter, so actions cannot be mapped to steps; both partly stem from our prompt. I think we can drop 34.3 and score unverifiable outputs as 0, e.g.\ all five verified: PDDL2.1 16.7\% vs.\ CP-SAT 93.9\%.}

% We additionally report \textbf{faithfulness diagnostics} that isolate formalization failures from solver failures.
% For AsyncPlan-XXL, where the ground-truth action graph is available, we compare the plan generated by Planner and formalization produced by Formalizer against the ground-truth actions $\mathcal{A}$, durations $\mathcal{T}$,
% precedence edges $\mathcal{C}_A$, and goals $g$ along five
% dimensions, detailed in Table~\ref{tab:diagnostic_metrics}.

\paragraph{Models and protocols.}
We evaluate four LLMs spanning proprietary and open-weight families: Gemini-3-Flash, GPT-5-mini, DeepSeek-V4-Flash, and Qwen3.6-35B-A3B, which we write as Qwen3.6-35B hereafter.
All models are run zero-shot at temperature $0$, following the protocol of~\citet{10.5555/3692070.3693283}. 
A stronger model, GPT-5.6 Sol, and a few-shot experiment are described in Appendix~\ref{app:strong_model}.
Prompts are provided in Appendix~\ref{app:prompts}.

\section{Results and Discussion}
\label{sec:results_discussion}

We organize the analysis as three progressive questions: whether LLMs can plan asynchronously at all (\S\ref{sec:rq1}), what drives the differences across target languages (\S\ref{sec:rq2}), and whether plans can be adapted as tasks change during execution (\S\ref{sec:rq3}).
% (Q1) On basic asynchronous planning, is the task much harder for Planner than for Formalizer?
% (Q2) Once we move beyond basic asynchronous planning into settings with high parallel speedup, online updates, and tight time limits, does Formalizer remain robust? And does the formal language (PDDL2.1 vs.\ CP-SAT) determine success, or does the LLM?

\subsection{Q1: Can LLMs plan asynchronously?}
\label{sec:rq1}

\begin{figure*}[t]
\centering
\includegraphics[width=0.7\linewidth]{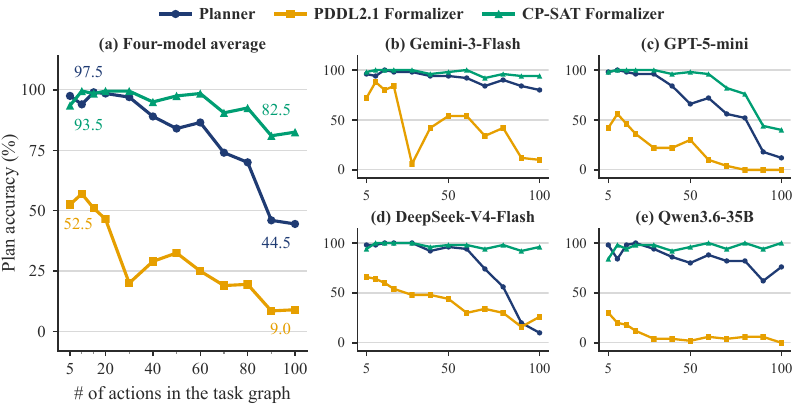}
\caption{\textbf{Plan accuracy on AsyncPlan-XXL vs. task-graph size.} Planner degrades as task graphs grow, while the CP-SAT \Formalizer remains robust. \textbf{(a)} Four-model average: from 5 to 100 actions, Planner drops from 97.5\% to 44.5\% and the CP-SAT \Formalizer from 93.5\% to 82.5\%; the PDDL2.1 \Formalizer is weakest throughout. \textbf{(b--e)} Per model: CP-SAT \Formalizer averages above 90\% for three of four models, while Planner at 100 actions varies from about 80\% to 10\%. Values and bootstrap intervals are in Tables~\ref{tab:asyncplanxxl_scaling_results} and~\ref{tab:uncertainty}.}
\label{fig:asyncplan_xxl_scaling}
\end{figure*}

\begin{figure*}[t]
\centering
\includegraphics[width=0.66\linewidth]{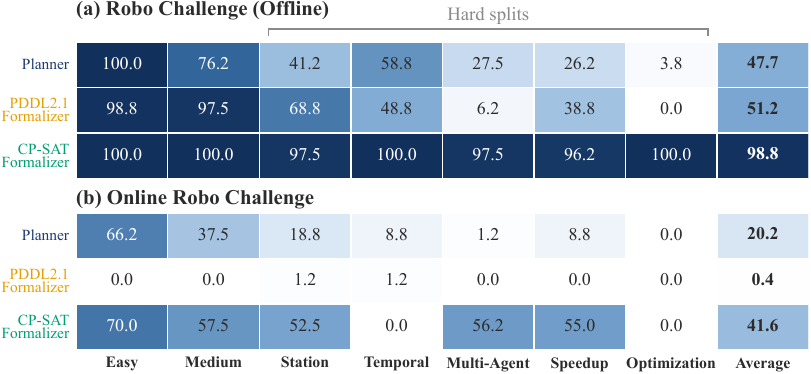}
\caption{\textbf{Plan accuracy by split on Robo Challenge, averaged over four LLMs.} The CP-SAT \Formalizer is near-perfect offline but degrades sharply online, while the other pipelines struggle on Hard splits even offline. \textbf{(a)} Offline: the CP-SAT \Formalizer solves nearly every task; Planner and the PDDL2.1 \Formalizer average about 50\%. \textbf{(b)} Online: all pipelines drop; the CP-SAT \Formalizer remains best but fails on Temporal and Optimization.
\vspace{-1em}
}
\label{fig:robo_splits}
\end{figure*}

\textbf{As tasks grow larger, the \cpsat stays robust while the \planner depends on the model.}
On the smallest tasks, those of AsyncHow and the 5-action tasks of AsyncPlan-XXL, the Planner and the CP-SAT \Formalizer both perform near ceiling and only the PDDL2.1 \Formalizer lags (Tables~\ref{tab:asyncplanxxl_scaling_results} and~\ref{tab:async_robo_full_results}). 
The gap opens as we scale task graphs of AsyncPlan-XXL to 100 actions. 
Averaged over four models, the Planner falls from 97.5\% at 5 actions to 44.5\% at 100, while the CP-SAT \Formalizer loses only about 10 points and the PDDL2.1 \Formalizer stays weakest (Figure~\ref{fig:asyncplan_xxl_scaling}(a)).  
This average hides large differences between models. At 100 actions the Planner stays above 75\% for Gemini-3-Flash and Qwen3.6-35B but drops to about 10\% for GPT-5-mini and DeepSeek-V4-Flash, while the CP-SAT \Formalizer stays above 90\% for three of the four models (Figure~\ref{fig:asyncplan_xxl_scaling}(b--e)).
Since every pipeline computes the makespan deterministically, the Planner's errors come mainly from the task graph it predicts. 
A stronger model closes the gap among all three pipelines. In Table~\ref{tab:strong_model_scaling}, GPT-5.6 Sol reaches 98.0\% with the Planner and the CP-SAT \Formalizer and 93.2\% with the PDDL2.1 \Formalizer, against four-model averages of 78.7\%, 93.9\%, and 27.5\% on the same five sizes. Yet CP-SAT already lifts weaker models such as Gemini-3-Flash and DeepSeek-V4-Flash to about 97\%, so we argue that a well-chosen formalization interface can substitute for a stronger model.

\begin{figure*}[t]
\centering
\includegraphics[width=0.62\linewidth]{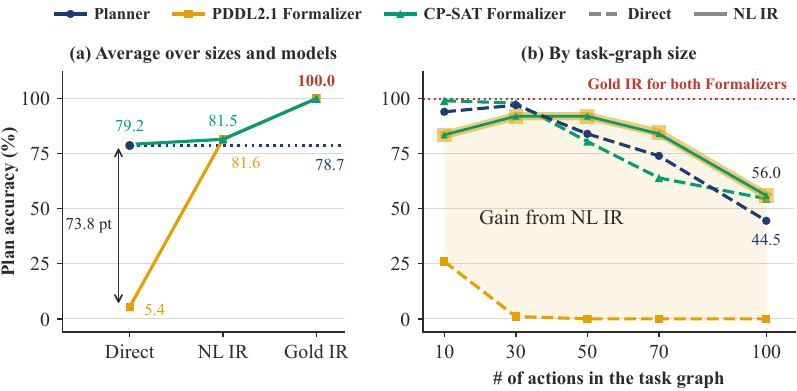}
\caption{\textbf{Plan accuracy of the Planner and \Formalizers under direct generation, NL IR and gold IR.} Colour identifies the pipeline and line style the condition, and all results use zero solver retries. \textbf{(a)} averages over sizes and models. Direct generation leaves PDDL2.1 far behind CP-SAT, but once the same model-extracted task graph is compiled into each language, both \Formalizers slightly outperform the Planner, and a gold graph lifts both to 100\%. \textbf{(b)} shows as task-graph size grows, the two NL IR curves nearly coincide. The shaded area is what NL IR gains for PDDL2.1, and the remaining decline with size comes from task-graph extraction rather than from either language.
\vspace{-1em}
}
\label{fig:ir_decomposition}
\end{figure*}

\paragraph{The same ordering holds under state constraints and widens online.}
On Robo Challenge, the CP-SAT \Formalizer is nearly perfect and stays above 96\% on every split, while the Planner and the PDDL2.1 \Formalizer fall to around 50\% and vary with the constraint type (Figure~\ref{fig:robo_splits}(a)); a paired task-cluster bootstrap finds a small but significant PDDL2.1 advantage over the Planner (Appendix~\ref{app:uncertainty}). 
Once execution-time events arrive, every pipeline degrades: CP-SAT drops to 41.6\%, the Planner to 20.2\%, and PDDL2.1 to 0.4\% (Figure~\ref{fig:robo_splits}(b)). Since the CP-SAT \Formalizer handles these tasks almost perfectly offline, the difficulty lies in keeping the specification current after each event (\S\ref{sec:rq3}).

% \S\ref{sec:rq3} asks whether updating the existing specification, rather than rewriting it, closes this gap.
% This gap includes generation, backend and execution errors. Section~\ref{sec:q2} evaluates the combined state-aware repair pipeline without attributing the entire difference to representation alone.

\paragraph{Hard splits reveal different pipeline weaknesses.} 
Across the Hard splits the Planner averages 3.8--58.8\%, while the PDDL2.1 \Formalizer reaches 68.8\% on \textit{Station} and 38.8\% on \textit{Speedup} but falls to 6.2\% on \textit{Multi-Agent} and 0\% on \textit{Optimization}; CP-SAT stays at or above 96.2\% on every offline split (Figure~\ref{fig:robo_splits}(a)). On \textit{Temporal} and \textit{Speedup}, one PDDL2.1 failure mode we observe is serialization through added predicate chains, although generation and backend errors also contribute.

\paragraph{\pddl fails where it should be strongest.}
PDDL2.1 was built for durative actions, precedence and concurrency, yet it is the weakest pipeline at scale and online, and on Robo Challenge it only narrowly beats the Planner. If expressiveness determined success the ordering would be reversed, so \S\ref{sec:rq2} asks whether the gap belongs to the formalism or to writing it.
% Taken together, these results show that LLMs can plan asynchronously on short, abstract instructions, but Planner collapses along three orthogonal axes: as the dependency graph grows, as action semantics become executable, and as the environment changes during execution. Across all three settings, the consistent finding is that pairing the LLM with a formal language whose primitives match the task structure, such as CP-SAT for state-constrained problems, Formalizer scales where direct generation does not.

\subsection{Q2: Does PDDL2.1 Fail Because of the Formalism or the Formalization?}
\label{sec:rq2}

% \begin{table}[t]
% \centering
% \small
% \begin{tabularx}{\linewidth}{@{}>{\raggedright\arraybackslash}p{0.40\linewidth}>{\raggedright\arraybackslash}X@{}}
% \toprule
% \textbf{Metric} & \textbf{Definition} \\
% \midrule
% \multicolumn{2}{@{}l}{\textit{Primary}} \\
% \textbf{Plan Accuracy} & $\pi$ satisfies all constraints in $\mathcal{C}$ and $T_{\mathrm{total}}(\pi) = T_{\mathrm{total}}^{\star}$. \\
% \midrule
% \multicolumn{2}{@{}l}{\textit{Faithfulness Diagnostics ($\mathcal{P}_A$)}} \\
% \textbf{Makespan Accuracy} & $T_{\mathrm{total}}(\pi) = T_{\mathrm{total}}^{\star}$. \\
% \textbf{Dependency Precision} & Fraction of predicted edges that appear in $\mathcal{C}_A$. \\
% \textbf{Dependency Recall} & Fraction of gold edges recovered. \\
% \textbf{Goal Completeness} & Fraction of $g$ encoded in the artifact. \\
% \textbf{Action Count Match} & $|\mathcal{A}_{\mathrm{pred}}| = |\mathcal{A}|$. \\
% \textbf{Duration Match} & $\forall a:\ \mathcal{T}_{\mathrm{pred}}(a) = \mathcal{T}(a)$. \\
% \bottomrule
% \end{tabularx}
% \caption{Evaluation metric for asynchronous planning. Plan accuracy is the primary  metrics across all settings; \emph{faithfulness diagnostics} aims to isolate formalization failures from solver failures for AsyncPlan-XXL.}
% \label{tab:diagnostic_metrics}
% \end{table}

We answer this question through two analyses. 
The faithfulness diagnostics first show where PDDL2.1 outputs go wrong. 
We then introduce an \textbf{intermediate representation (IR)}, a solver-neutral JSON description of the task graph (e.g. Figure~\ref{fig:ir_example}).
Asking the model for the IR rather than for the target language separates the formalism from the act of writing it. 
Both use AsyncPlan-XXL, which provides a gold task graph and reduces the remaining work to deterministic scheduling, so a pipeline's output can be compared against that graph directly.

\paragraph{The diagnostics suggest that \pddl fails in writing rather than in task understanding.}
Table~\ref{tab:faithfulness} verifies 93.9\% of CP-SAT specifications and 79.1\% of Planner outputs, but only 16.7\% of PDDL2.1 specifications. Many PDDL2.1 outputs fail before their content can be judged at all, with 34.7\% unparsable, mostly because of syntax errors. Every verified output also yields a correct plan, so the diagnostics capture what plan accuracy depends on. The converse does not hold, because a plan can be correct even when we cannot parse the output it came from, or when an extra precedence edge leaves the optimal schedule unchanged. PDDL2.1 therefore breaks down when the model writes the language, and the IR separates that cost from that of graph extraction.
% An output is verified when all five diagnostics can be computed and give full marks. On AsyncPlan-XXL, 93.9\% of CP-SAT specifications and 79.1\% of Planner outputs are verified, but only 16.7\% of PDDL2.1 specifications are (Table~\ref{tab:faithfulness}). Many unverified PDDL2.1 outputs fail before their content can be judged: 34.7\% cannot be parsed at all, mostly due to syntax errors. Every verified record also has a correct plan, so the diagnostics capture what plan accuracy depends on; the converse need not hold, since an accurate plan may have an unreadable structure or an extra precedence edge off the critical path. These results suggest that PDDL2.1 breaks down when the model writes the language, which the IR control further separates into the cost of writing the target language and the cost of extracting the task graph.

\paragraph{The IR locates the gap in formalization.} We compare three conditions. Under \textbf{Direct}, the model writes the formal specification itself. Under \textbf{NL IR}, it writes only the IR, so the gain over Direct measures the cost of writing the target language. Under \textbf{Gold IR}, we give the task graph, so the gap between NL IR and Gold IR measures the cost of extracting it from the task description.
Under Direct, PDDL2.1 trails CP-SAT by 73.8 points and solves no task with 50 or more actions (Figure~\ref{fig:ir_decomposition}(a, b)). 
Once the model only extracts the graph, PDDL2.1 gains more than 75 points and the gap to CP-SAT shrinks to 0.1 points (Figure~\ref{fig:ir_decomposition}(a)), so the model understands most of the tasks but fails to express them correctly in PDDL2.1 directly. Given the gold graph, both target languages reach 100\% (Figure~\ref{fig:ir_decomposition}(a)), and the independent PDDL2.1 temporal planner POPF~\citep{coles2010popf} also solves every compiled problem, so in this precedence-only setting neither the language nor the solver is the obstacle, as Appendix~\ref{app:ir_compilation} details. 
% We compare Direct generation, where the model writes the formal specification itself, with the IR, which a deterministic compiler turns into PDDL2.1 or CP-SAT code for the solver. In \textbf{NL IR}, the model only extracts the graph, so the gain over Direct measures the cost of writing the target language. In \textbf{Gold IR}, the graph is given, so the gap between NL IR and Gold IR measures the cost of graph extraction.
% Under Direct generation, PDDL2.1 trails CP-SAT by 73.8 points (Figure~\ref{fig:ir_decomposition}(a)) and solves no task with 50 or more actions (Figure~\ref{fig:ir_decomposition}(b)). Once the model only extracts the graph, PDDL2.1 gains more than 75 points and the gap to CP-SAT shrinks to 0.1 points (Figure~\ref{fig:ir_decomposition}(a)), so the model understands most of the tasks but fails to express them correctly in PDDL2.1 directly. Given the gold graph, both target languages reach 100\% (Figure~\ref{fig:ir_decomposition}(a)), and the independent PDDL2.1 temporal planner POPF~\citep{coles2010popf} also solves every compiled problem, so in this precedence-only setting neither the language nor the solver is the obstacle, as Appendix~\ref{app:ir_compilation} details. 

\paragraph{Graph extraction remains the shared bottleneck, which retries and stronger models ease but do not remove.}
Once the compiler and solvers are deterministic and the gold graph yields a perfect ceiling, whatever accuracy NL IR still loses must come from extracting the task graph. The Planner, which turns its own graph into a schedule by longest-path arithmetic, performs similarly to both NL IR pipelines, and all three decline from above 80\% at 10 actions and end at 100 actions with 56.0\% for the two NL IR pipelines and 44.5\% for the Planner (Figure~\ref{fig:ir_decomposition}(b)). 
Extraction is therefore a bottleneck every pipeline shares, which the IR locates rather than removes.
A stronger model narrows the gap. Even with up to three retries, PDDL2.1 stays more than 60 points behind CP-SAT, as Table~\ref{tab:common_ir_control_full} in Appendix~\ref{app:common_ir_control} shows, whereas a stronger generator, GPT-5.6 Sol, narrows the gap to 4.8 points (\S\ref{sec:rq1}). Few-shot exemplars do not help consistently (Appendix~\ref{app:fewshot}).

\subsection{Q3: Can LLMs Adapt Plans as Tasks Change During Execution?}
\label{sec:rq3}

\begin{figure}[t]
\centering
\includegraphics[width=0.78\columnwidth]{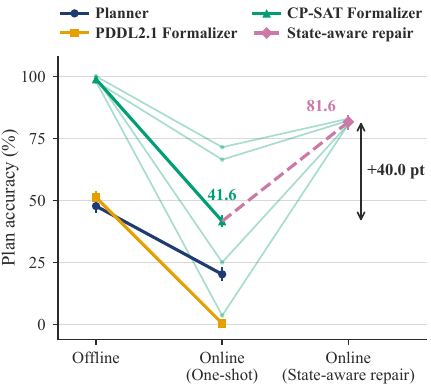}
\caption{\textbf{Average plan accuracy on Robo Challenge offline and online, with one-shot re-formalization and with state-aware CP-SAT repair.} Faint lines show individual models. Updating the specification after each event, instead of rewriting it, raises CP-SAT from 41.6\% to 81.6\% and removes most of the spread between models.
\vspace{-1em}
}
\label{fig:state-aware}
\end{figure}

% PDDL2.1 fails wherever the LLM must maintain an extra symbolic layer consistently.
\S\ref{sec:rq1} showed that even the CP-SAT \Formalizer loses more than half of its accuracy once events arrive. We therefore compare two ways of handling an event, both of which must keep the already-executed prefix intact. 
\emph{One-shot re-formalization} rewrites the whole specification from the revised task description. \emph{State-aware repair} gives the model the event and the current execution state, asks only for changes to the affected constraints, and lets a deterministic compiler rebuild the rest.

\paragraph{Updating the affected constraints doubles online accuracy and makes it stable across models.}
With one-shot re-formalization the CP-SAT \Formalizer solves 41.6\% of online tasks, whereas state-aware repair solves 81.6\%, with a bootstrap interval from 78.8\% to 84.1\%, as Figure~\ref{fig:state-aware} shows. 
Repair is also far more stable. 
One-shot accuracy ranges from almost nothing to about 70\% across the four models, while repair stays within about two points for all of them, so updating rather than rebuilding also makes the result far less model-dependent. Nor is event interpretation the obstacle, since the applied constraint changes matched the ground-truth update in all 1,475 repair steps reached, excluding steps after an episode had already failed. 
% \jiayi{need more implication}
This suggests that online failures stem less from reasoning about the event than from rewriting the entire specification at every event: each rewrite is a fresh chance to drop committed state or reintroduce errors, echoing \S\ref{sec:rq2}'s finding that writing the formal language, not understanding the task, is the main source of error. A formal specification is thus valuable online not only because a solver can check it, but because it can be edited locally, so online planners should keep the specification as persistent state and ask the LLM only for the changes an event implies. 
Because repair also removes most of the spread between models, the choice of interface matters more here than the choice of model.

\section{Conclusion}
\label{sec:conclusion}
We systematically studied LLMs as Planners and as \Formalizers on asynchronous planning, across action-constrained, state-constrained, and online settings, on three benchmarks we release. The CP-SAT \Formalizer was the most robust to task complexity, the Planner's robustness depended on the model, and PDDL2.1, the formalism built for temporal planning, was the weakest. Our controls show that this weakness lies in writing the formalism, not in the formalism itself. When both specifications are compiled from the same model-extracted task graph, the 73.8-point gap under direct generation shrinks to 0.1 points, and the gold graph lifts both to 100\%. What remains is a bottleneck shared by every pipeline, recovering the task graph from language, which solver feedback and stronger models ease but do not remove. The same pattern holds online, where updating only the event-induced constraints reaches 81.6\% against 41.6\% for rewriting the whole specification. Overall, how well the interface matches the task structure matters more than how expressive the target language is, and a well-chosen interface can let a weaker model match a stronger one.

\newpage
\section*{Limitations}

\textbf{Benchmark scope.} Our benchmarks are controlled rather than naturalistic. AsyncPlan-XXL is synthetic, and we generate Robo Challenge and Online Robo Challenge procedurally as cooking tasks with clean identifiers. 
This control lets us raise difficulty step by step, from durations and precedence to state and resource constraints to execution-time events, so we can test each ability that asynchronous planning requires and trace errors to their source during analysis, but it also means our findings concern scheduling-structured asynchronous planning, not messier natural language, richer embodied environments, or domains such as logistics and rail dispatch~\cite{mohanty2020flatlandrlmultiagentreinforcement}. For the same reason, strong results here do not make these pipelines ready for deployment: real robotics, logistics, and dispatch systems impose safety and real-time requirements that we do not evaluate, so generated plans need domain validation and human oversight before anyone executes them.

\textbf{Pipeline and analysis scope.} Our conclusions about formalisms also rest on a narrow set of comparisons. We pair PDDL2.1 with OPTIC, which is widely used for temporal planning, and use CP-SAT scheduling JSON as a representative scheduling-native target, which differs in expressiveness and is not necessarily the optimal choice. Other formalisms (PDDL3, ANML, SMT, MiniZinc) or solvers (LPG, TFD) could shift the trade-off, and we test POPF only on the gold-structure control (Appendix~\ref{app:ir_compilation}). Our controls are similarly limited: we run the IR analysis only on AsyncPlan-XXL, where a gold task graph exists, and state-aware repair only for CP-SAT, so neither yet tests how PDDL2.1 handles resource and state constraints.

\textbf{Models and statistical power.} Finally, our model coverage and sample sizes limit how far the numbers generalize. We evaluate four models zero-shot at temperature 0, and the one stronger model we test, GPT-5.6 Sol, largely closes the gaps between pipelines on AsyncPlan-XXL. Each Robo Challenge split has only 20 tasks, a size we chose to keep human-verified ground truth tractable, so we treat split-level differences as trends rather than precise effect sizes; Appendix~\ref{app:uncertainty} reports bootstrap confidence intervals for these comparisons. Whether the pipeline ordering and the offline-to-online gap hold with stronger models or larger benchmarks remains open.
% \jiayi{limited data size of robo challenge, and mention that we have test on larger scale with open source model and confirm the trend is similar, any way to prove that the current scale is enough to get the statiscal significant conclusion for our research question?}

% \section*{Acknowledgments}

\bibliography{custom}

\newpage
\appendix
\renewcommand{\topfraction}{0.95}\renewcommand{\bottomfraction}{0.9}\renewcommand{\textfraction}{0.05}\renewcommand{\floatpagefraction}{0.85}\setcounter{topnumber}{4}\setcounter{totalnumber}{6}

This appendix is organized as follows. Section~\ref{app:benchmarks} describes the benchmarks (statistics, benchmark verification and example data), followed by potential risks in Section~\ref{app:potential risks}. Section~\ref{app:detailed_results} collects the extended results: scaling, Robo Challenge, the controlled IR analysis, stronger models and few-shot prompting. Section~\ref{app:eval_protocol} gives the evaluation details, the error analysis (including the Robotouille failure modes) and uncertainty estimates. Section~\ref{app:qualitative_errors} shows qualitative error examples, and Section~\ref{app:prompts} lists all prompts.

\section{Benchmarks and Statistics}
\label{app:benchmarks}

\subsection{Benchmarks Statistics}
\label{app:dataset_stats}

This section describes the benchmarks used in our experiments.
We evaluate asynchronous planning under three settings: action-constrained planning using AsyncHow~\cite{10.5555/3692070.3693283} and AsyncPlan-XXL, state-constrained planning using Robotouille~\cite{gonzalez-pumariega2025robotouille} and Robo Challenge, and online planning using Online Robo Challenge.
Both datasets we build on are distributed under the MIT License, which permits our modification and redistribution.
Table~\ref{tab:synth_dataset_stats} reports the overall scale and evaluation focus of these benchmarks.
Statistics for the action-constrained benchmark are provided in Table~\ref{tab:prev_dataset_stats}, while Table~\ref{tab:robo-challenge-stats} presents statistics for the state-constrained and online benchmarks.

% preamble: \usepackage{booktabs, tabularx, array}
\begin{table*}[!htbp]
\centering
\footnotesize
\renewcommand{\arraystretch}{1.25}
\setlength{\tabcolsep}{5pt}
\newcolumntype{Y}{>{\raggedright\arraybackslash}X}
\begin{tabularx}{\linewidth}{@{}l l c l l Y@{}}
\toprule
\textbf{Benchmark} & \textbf{Setting} & \textbf{Inst.} & \textbf{Avg.\ size} & \textbf{Stressed axis} & \textbf{Evaluation focus} \\
\midrule
AsyncHow              & Action-constrained & 320 & 5.14 steps         & \textemdash           & Duration/dependency extraction \\
AsyncPlan-XXL         & Action-constrained & 600 & 47.50 steps        & Structural complexity & Action-graph scaling \\
\cmidrule{1-6}
Robo Challenge        & State-constrained  & 140 & 8.53 items         & Constraint density    & Grounded robot constraints \\
Online Robo Challenge & Online             & 140 & 8.53 initial items & Temporal openness     & Dynamic replanning \\
\bottomrule
\end{tabularx}
\caption{\textbf{Summary of evaluation benchmarks.} Each setting is anchored by an existing benchmark and extended by a new benchmark that stresses one axis of difficulty.}
\label{tab:synth_dataset_stats}
\end{table*}

\begin{table*}[t]
\centering
\small
\setlength{\tabcolsep}{6pt}
\begin{tabular}{llrccccc}
\toprule
\textbf{Benchmark} & \textbf{Subset} & \textbf{\#Instances} & \textbf{\#Steps} & \textbf{\#Edges} & \textbf{CP} & \textbf{Par. Ratio} & \textbf{AND-Joins} \\
\midrule
AsyncHow      & --      & 320 & 3--17 (5.1)  & 1--19 (4.2) & 3.58  & --   & --   \\
Robotouille   & --      &  10 & 19--82 (43)  & --          & --    & --   & --   \\
\midrule
\multirow{12}{*}{AsyncPlan-XXL}
              & $N{=}5$    &  50 &   5 & $3.7 \pm 0.7$    &  3.26 & 1.29 &  0.3 \\
              & $N{=}10$   &  50 &  10 & $8.8 \pm 1.2$    &  5.74 & 1.53 &  1.2 \\
              & $N{=}15$   &  50 &  15 & $14.8 \pm 1.7$   &  6.62 & 1.58 &  2.8 \\
              & $N{=}20$   &  50 &  20 & $19.8 \pm 2.3$   &  7.04 & 2.02 &  4.0 \\
              & $N{=}30$   &  50 &  30 & $35.9 \pm 4.3$   &  8.58 & 2.14 &  9.2 \\
              & $N{=}40$   &  50 &  40 & $54.8 \pm 5.8$   &  9.08 & 2.05 & 15.1 \\
              & $N{=}50$   &  50 &  50 & $78.5 \pm 6.6$   &  9.56 & 2.87 & 21.8 \\
              & $N{=}60$   &  50 &  60 & $104.7 \pm 9.5$  & 10.82 & 3.34 & 28.8 \\
              & $N{=}70$   &  50 &  70 & $142.4 \pm 10.6$ & 12.00 & 3.16 & 38.5 \\
              & $N{=}80$   &  50 &  80 & $179.6 \pm 12.5$ & 13.40 & 3.17 & 46.4 \\
              & $N{=}90$   &  50 &  90 & $222.0 \pm 11.1$ & 14.34 & 3.28 & 56.3 \\
              & $N{=}100$  &  50 & 100 & $271.1 \pm 17.7$ & 15.62 & 3.01 & 66.2 \\
\cmidrule(l){2-8}
              & Total      & 600 & 5--100 & --             & 3.26--15.62 & 1.29--3.34 & 0.3--66.2 \\
\bottomrule
\end{tabular}
\caption{\textbf{Benchmark statistics for the AsyncHow, Robotouille and AsyncPlan-XXL benchmarks.} For AsyncHow and Robotouille, \#Steps and \#Edges are reported as min--max (mean); for AsyncPlan-XXL, each row corresponds to one synthetic graph size and \#Edges is reported as mean $\pm$ std. CP is the average critical-path length (number of nodes on the longest precedence chain). Par.\ Ratio is the sequential-to-optimal makespan ratio (total sequential duration / critical-path duration); AND-Joins counts nodes with $\geq$2 predecessors. 
% \jiayi{fill in std and AND-Join counts for $N{=}60,70,80,90$, check numbers}\cathy{added}
}
\label{tab:prev_dataset_stats}
\end{table*}

\begin{table}[!htbp]
\centering
\scriptsize
\begin{tabularx}{\linewidth}{@{}>{\raggedright\arraybackslash}Xrrrrr@{}}
\toprule
\textbf{Split} & \makecell{\textbf{Inst.}} & \makecell{\textbf{Avg.}\\\textbf{Items}} & \makecell{\textbf{Avg.}\\\textbf{Opt.}} & \makecell{\textbf{Avg.}\\\textbf{Speedup}} & \makecell{\textbf{Online}\\\textbf{Events}} \\
\midrule
Easy & 20 & 2.70 & 13.20 & 1.24 & 2 \\
Medium & 20 & 5.00 & 18.75 & 1.84 & 2 \\
Hard High-Speedup & 20 & 17.00 & 24.00 & 6.88 & 3 \\
Hard Multi-Agent & 20 & 10.00 & 39.10 & 2.33 & 3 \\
Hard Optimization & 20 & 7.00 & 24.75 & 2.00 & 3 \\
Hard Station & 20 & 10.00 & 25.00 & 3.52 & 3 \\
Hard Temporal & 20 & 8.00 & 29.00 & 2.22 & 3 \\
\bottomrule
\end{tabularx}
\caption{\textbf{Robo Challenge and Online Robo Challenge statistics by split.} Avg. Opt. is the average optimal makespan computed by the CP-SAT oracle, and Avg. Speedup is the ratio between sequential and optimal makespan. Each Online Robo Challenge episode starts from the Robo Challenge task of the same split (same instance count and items) and adds the number of execution-time events shown, such as new deliveries, deadlines, or resource changes.
\vspace{-1em}
}
\label{tab:robo-challenge-stats}
\end{table}

\subsection{Example Data}
\label{app:example_data}
% B4 Data Contains Personally Identifying Info Or Offensive Content*
% Did you discuss the steps taken to check whether the data that was collected/used contains any information that names or uniquely identifies individual people or offensive content, and the steps taken to protect/anonymize it?
% B4 Elaboration
% [COMPULSORY IF YES/NO] For yes, provide a section number. For no, justify why not.
We verified that none of our benchmarks contain personally identifying information or offensive content.
AsyncHow is derived from publicly available WikiHow procedural articles describing generic everyday tasks~\citep{10.5555/3692070.3693283}.
Robotouille consists of procedurally generated cooking scenarios~\citep{gonzalez-pumariega2025robotouille}, which we transform from the original JSON-formatted data points into natural language for readability.
AsyncPlan-XXL is fully synthetic, with DAG structures randomly generated and rewritten into natural language by Gemini-3-Flash.
Robo Challenge and Online Robo Challenge are also procedurally generated from symbolic cooking tasks; they contain artificial item names, station capacities, action durations, deliveries, and event streams rather than human subjects or real-world personal records.
We manually inspected samples across all five benchmarks and confirmed that no personally identifying information, offensive content, or sensitive data is present.
Randomly selected examples from each benchmark are shown in~\Cref{tab:example_data_point,tab:example_data_point_cont}.

\begin{table*}[t]
\centering
\scriptsize
\renewcommand{\arraystretch}{0.8}
\begin{tabular}{p{2.4cm} p{12.3cm}}
\toprule
\textbf{Benchmark} & \textbf{Example Problem} \\
\midrule
AsyncHow &
To go to the local water park, here are the steps and the times needed for each step.
\newline Step 1. Put on swim suit (15 minutes)
\newline Step 2. Search online for local water park (1 hour)
\newline Step 3. Find out pricing and dates it is open (15 minutes)
\newline Step 4. Purchase tickets to park (5 minutes)
\newline Step 5. Get into vehicle (5 minutes)
\newline
\newline These ordering constraints need to be obeyed when executing the above steps:
\newline Step 1 must precede Step 5.
\newline Step 2 must precede Step 3.
\newline Step 3 must precede Step 4.
\newline Step 4 must precede Step 5.
\newline
\newline Question: Assume that you need to execute all the steps to complete the task and that infinite resources are available. What is the shortest possible time to go to the local water park? \\
\midrule
AsyncPlan-XXL &
To buy postage stamps, here are the steps needed.
\newline Step 1. Check current stamp inventory at home (5 minutes)
\newline Step 2. Walk from the parking lot into the post office (3 minutes)
\newline Step 3. Select the specific stamp design from the display (2 minutes)
\newline Step 4. Find the car keys (1 minute)
\newline Step 5. Research the nearest post office location online (10 minutes)
\newline Step 6. Grab a wallet and identification (2 minutes)
\newline Step 7. Drive to the post office location (20 minutes)
\newline Step 8. Wait in the queue for an available teller (15 minutes)
\newline Step 9. Drive to the post office area (15 minutes)
\newline Step 10. Request the desired number of stamp coils (1 minute)
\newline Step 11. Create a shopping list for office supplies (5 minutes)
\newline Step 12. Locate a nearby ATM to withdraw cash (10 minutes)
\newline Step 13. Park the car in the visitor lot (4 minutes)
\newline Step 14. Check the post office hours on the door (30 seconds)
\newline Step 15. Pay the teller for the stamps (2 minutes)
\newline
\newline These ordering constraints need to be obeyed when executing the above steps:
\newline Step 1 must precede Step 11.
\newline Step 2 must precede Step 10.
\newline Step 2 must precede Step 15.
\newline Step 3 must precede Step 8.
\newline Step 4 must precede Step 9.
\newline Step 5 must precede Step 7.
\newline Step 6 must precede Step 9.
\newline Step 7 must precede Step 8.
\newline Step 7 must precede Step 13.
\newline Step 9 must precede Step 3.
\newline Step 9 must precede Step 12.
\newline Step 11 must precede Step 2.
\newline Step 13 must precede Step 14.
\newline Step 14 must precede Step 2.
\newline
\newline Question: Assume that you need to execute all the steps to complete the task and that infinite resources are available. What is the shortest possible time to buy postage stamps? \\
\bottomrule
\end{tabular}
\caption{\textbf{Example questions or task descriptions from the action-constrained benchmarks.}}
\label{tab:example_data_point}
\end{table*}

\begin{table*}[t]
\centering
\scriptsize
\renewcommand{\arraystretch}{0.8}
\begin{tabular}{p{2.4cm} p{12.3cm}}
\toprule
\textbf{Benchmark} & \textbf{Example Problem} \\
\midrule
Robotouille &
You are a robot in a kitchen environment. You are currently in a 5x5 kitchen at your starting position facing up.
\newline
\newline There are 4 stations: table\_1, stove\_1, table\_2, and table\_3.
\newline There are 4 items: bread\_1 on table\_1, bread\_2 on table\_1, chicken\_1 on table\_2, and cheese\_1 on table\_3.
\newline Here are some constraints on the items: chicken\_1 must be cooked, which will take 3 steps on a stove. bread\_1, chicken\_1, and cheese\_1 must be stacked before bread\_2.
\newline
\newline Now you need to prepare a cheese chicken sandwich on a table. To do so, make sure: bread\_1 is on table, chicken is cooked, chicken is at table, cheese is at table, bread\_2 is at table, and bread\_2 has nothing on top of it. \\
\midrule
Robo Challenge &
Task \texttt{easy\_01\_grilled\_chicken\_sandwich} starts with three items: \texttt{chicken} is raw and cookable, while \texttt{bun\_bot} and \texttt{bun\_top} are ready.
\newline
\newline The goal is to grill the chicken and stack the sandwich in the order \texttt{bun\_bot}, \texttt{chicken}, \texttt{bun\_top}.
\newline Available station capacities are: grill 1, cutting board 1, fryer 1, boiler 1, toaster 1, and marinator 1.
\newline Action durations include grill 10 seconds, cut 3 seconds, fry 8 seconds, boil 12 seconds, toast 5 seconds, marinate 6 seconds, mash 4 seconds, and stack 1 second.
\newline
\newline Question: Produce a valid schedule that minimizes the makespan while respecting action preconditions, station capacities, and the required stack order. \\
\midrule
Online Robo Challenge &
Episode \texttt{online\_easy\_01\_grilled\_chicken\_sandwich} begins from the Robo Challenge grilled-chicken-sandwich task.
\newline
\newline Initial task: grill \texttt{chicken} and stack \texttt{bun\_bot}, \texttt{chicken}, \texttt{bun\_top}; the source optimal makespan is 12 seconds.
\newline Event 1 at $t{=}2.29$ seconds: a new order arrives requiring \texttt{online\_cut\_01\_11} to be cut.
\newline Event 2 at $t{=}5.17$ seconds: another order arrives requiring \texttt{online\_grill\_01\_21} to be grilled.
\newline
\newline Question: Continue execution after each event by preserving completed and ongoing actions, adding the newly revealed work, and repairing the remaining schedule under the same station capacities and action durations. \\
\bottomrule
\end{tabular}
\caption{\textbf{Example questions or task descriptions from the state-constrained and online benchmarks.}}
\label{tab:example_data_point_cont}
\end{table*}

\subsection{Benchmark Verification}
\label{app:benchmark_verification}

\paragraph{AsyncPlan-XXL rewriting.}
The second stage of the generation pipeline rewrites an abstract DAG into a natural-language planning problem (Figure~\ref{fig:nl_generation_prompt}), which risks introducing constraints that the gold graph does not contain.
A validator therefore accepts a rewritten instance only if it preserves the step count exactly and if its set of explicit precedence edges is exactly equal to that of the source DAG.
All 600 instances pass this check.
We additionally spot-checked instances by hand across all twelve graph sizes and found no unintended ordering language introduced by the rewrite.

\paragraph{Robo Challenge oracle.}
Benchmark construction retains an instance only when its oracle solve reports \textsc{Optimal} and its witness passes the feasibility validator. All 140 stored witnesses pass the duration, execution, temporal and goal checks, and their makespans match the recorded values. Optimization witnesses also attain the recorded reward within inventory and deadline constraints. This check covers the witnesses; optimality rests on the oracle status recorded at construction, not on an independent solver run.

\paragraph{Online continuations.}
For Online Robo Challenge the implementation guarantees that the continuation it constructs is feasible, rather than proving that every event-conditioned continuation admits a feasible solution.
Across 3{,}610 continuation solves, 3{,}410 returned \textsc{Optimal} and 200 returned \textsc{No\_Feasible\_Subset}; no timed-out or unproven solution was accepted into the gold data.

\section{Potential Risks}
\label{app:potential risks}
% A2 Potential Risks*
% Did you discuss any potential risks of your work?
% Yes
% A2 Elaboration
% [COMPULSORY IF YES/NO] For yes, provide a section number. For no, justify why not.
All evaluated pipelines can return incorrect or incomplete plans. A matching makespan does not establish action coverage, correct durations or feasibility under true execution constraints. Tables~\ref{tab:faithfulness} and~\ref{tab:stepwise-diagnostics} distinguish supported specification diagnostics from end-to-end schedule validation. Online evaluation further checks event arrival times, non-cancellable actions, temporal lags and deadlines. These controlled evaluations do not establish deployment safety; generated plans need validation against the actual operating environment.

The Robo Challenge and Online Robo Challenge benchmarks introduce executable action semantics, station capacities, multi-agent assignment, deadlines, and execution-time events. While these settings move closer to realistic agentic deployment, they remain abstractions of full embodied systems: we do not model continuous dynamics, partial observability, sensor noise, or adversarial conditions. Plans that succeed in our evaluation should not be interpreted as ready for execution on physical robots without additional safety layers.

Finally, our AsyncPlan-XXL benchmark uses Gemini-3-Flash to rewrite abstract DAG structures into natural language descriptions of planning problems. This introduces a dependency on a commercial LLM during data construction, which may embed subtle biases in task framing, duration assignments, or activity naming. To mitigate this and enable reproducibility independent of any specific LLM, we release the underlying DAG structures alongside the generated natural language descriptions, so future work can re-render the same structural instances with different language models or human writers.

\section{Extended Results}
\label{app:detailed_results}

\subsection{Additional Results Across Benchmarks}

Table~\ref{tab:asyncplanxxl_scaling_results} reports stepwise plan accuracy (a)
and makespan accuracy (b) on AsyncPlan-XXL for each model and pipeline: Planner,
PDDL2.1 Formalizer, and CP-SAT Formalizer.

Table~\ref{tab:async_robo_full_results} reports per-model makespan accuracy on AsyncHow~\cite{10.5555/3692070.3693283} and simulator execution success on Robotouille~\cite{gonzalez-pumariega2025robotouille}. AsyncHow measures objective agreement, not full schedule feasibility; PDDL makespans are computed from returned schedules. Robotouille's metric, task count and configuration appear in Appendix~\ref{app:robotouille}.
Tables~\ref{tab:robo_challenge_per_split} and~\ref{tab:robo_challenge_combined} give split-level and per-model results for Robo Challenge, Online Robo Challenge and state-aware repair. Repair raises average online CP-SAT plan accuracy from 41.6\% to 81.6\%.

% preamble: \usepackage[table]{xcolor}  \usepackage{booktabs}
% (load xcolor with the table option, which also loads colortbl for \rowcolor)
\begin{table*}[t]
\centering\scriptsize
\renewcommand{\arraystretch}{1.12}
\setlength{\tabcolsep}{3pt}
\begin{tabular}{@{}l *{12}{r} @{\hskip 9pt} r@{}}
\toprule
\diagbox[width=7em]{\textbf{Model}}{\textbf{\# Actions}} & \textbf{5} & \textbf{10} & \textbf{15} & \textbf{20} & \textbf{30} & \textbf{40} & \textbf{50} & \textbf{60} & \textbf{70} & \textbf{80} & \textbf{90} & \textbf{100} & \textbf{Avg.} \\
\midrule
\multicolumn{14}{@{}l}{\textbf{(a) Plan accuracy (\%)}} \\
\midrule
\planner \\
Gemini-3-Flash & 96.0 & 94.0 & 100.0 & 98.0 & 98.0 & 94.0 & 94.0 & 92.0 & 84.0 & 90.0 & 84.0 & 80.0 & 92.0 \\
GPT-5-mini & 98.0 & 100.0 & 98.0 & 96.0 & 96.0 & 84.0 & 66.0 & 72.0 & 56.0 & 52.0 & 18.0 & 12.0 & 70.7 \\
DeepSeek-V4-Flash & 98.0 & 98.0 & 100.0 & 100.0 & 100.0 & 92.0 & 96.0 & 94.0 & 74.0 & 56.0 & 20.0 & 10.0 & 78.2 \\
Qwen3.6-35B & 98.0 & 84.0 & 98.0 & 100.0 & 94.0 & 86.0 & 80.0 & 88.0 & 82.0 & 82.0 & 62.0 & 76.0 & 85.8 \\
\avg & 97.5 & 94.0 & 99.0 & 98.5 & 97.0 & 89.0 & 84.0 & 86.5 & 74.0 & 70.0 & 46.0 & 44.5 & 81.7 \\
\addlinespace[3pt]
\pddl \\
Gemini-3-Flash & 72.0 & 88.0 & 80.0 & 84.0 & 6.0 & 42.0 & 54.0 & 54.0 & 34.0 & 42.0 & 12.0 & 10.0 & 48.2 \\
GPT-5-mini & 42.0 & 56.0 & 46.0 & 36.0 & 22.0 & 22.0 & 30.0 & 10.0 & 4.0 & 0.0 & 0.0 & 0.0 & 22.3 \\
DeepSeek-V4-Flash & 66.0 & 64.0 & 60.0 & 54.0 & 48.0 & 48.0 & 44.0 & 30.0 & 34.0 & 30.0 & 16.0 & 26.0 & 43.3 \\
Qwen3.6-35B & 30.0 & 20.0 & 18.0 & 12.0 & 4.0 & 4.0 & 2.0 & 6.0 & 4.0 & 6.0 & 6.0 & 0.0 & 9.3 \\
\avg & 52.5 & 57.0 & 51.0 & 46.5 & 20.0 & 29.0 & 32.5 & 25.0 & 19.0 & 19.5 & 8.5 & 9.0 & 30.8 \\
\addlinespace[3pt]
\cpsat \\
Gemini-3-Flash & 98.0 & 100.0 & 100.0 & 100.0 & 100.0 & 96.0 & 98.0 & 100.0 & 92.0 & 96.0 & 94.0 & 94.0 & 97.3 \\
GPT-5-mini & 98.0 & 100.0 & 100.0 & 100.0 & 100.0 & 96.0 & 98.0 & 96.0 & 82.0 & 76.0 & 44.0 & 40.0 & 85.8 \\
DeepSeek-V4-Flash & 94.0 & 100.0 & 100.0 & 100.0 & 100.0 & 96.0 & 98.0 & 98.0 & 94.0 & 98.0 & 92.0 & 96.0 & 97.2 \\
Qwen3.6-35B & 84.0 & 98.0 & 94.0 & 98.0 & 98.0 & 92.0 & 96.0 & 100.0 & 94.0 & 100.0 & 94.0 & 100.0 & 95.7 \\
\avg & 93.5 & 99.5 & 98.5 & 99.5 & 99.5 & 95.0 & 97.5 & 98.5 & 90.5 & 92.5 & 81.0 & 82.5 & 94.0 \\
\midrule[\heavyrulewidth]
\multicolumn{14}{@{}l}{\textbf{(b) Makespan accuracy (\%)}} \\
\midrule
\planner \\
Gemini-3-Flash & 96.0 & 94.0 & 100.0 & 98.0 & 98.0 & 96.0 & 94.0 & 96.0 & 90.0 & 94.0 & 88.0 & 84.0 & 94.0 \\
GPT-5-mini & 98.0 & 100.0 & 100.0 & 96.0 & 98.0 & 90.0 & 76.0 & 86.0 & 82.0 & 78.0 & 58.0 & 50.0 & 84.3 \\
DeepSeek-V4-Flash & 98.0 & 100.0 & 100.0 & 100.0 & 100.0 & 98.0 & 98.0 & 98.0 & 78.0 & 62.0 & 20.0 & 16.0 & 80.7 \\
Qwen3.6-35B & 98.0 & 84.0 & 98.0 & 100.0 & 94.0 & 90.0 & 88.0 & 88.0 & 84.0 & 90.0 & 68.0 & 80.0 & 88.5 \\
\avg & 97.5 & 94.5 & 99.5 & 98.5 & 97.5 & 93.5 & 89.0 & 92.0 & 83.5 & 81.0 & 58.5 & 57.5 & 86.9 \\
\addlinespace[3pt]
\pddl \\
Gemini-3-Flash & 76.0 & 96.0 & 86.0 & 92.0 & 6.0 & 44.0 & 64.0 & 56.0 & 38.0 & 52.0 & 26.0 & 26.0 & 55.2 \\
GPT-5-mini & 70.0 & 66.0 & 56.0 & 40.0 & 26.0 & 34.0 & 50.0 & 38.0 & 20.0 & 24.0 & 10.0 & 10.0 & 37.0 \\
DeepSeek-V4-Flash & 82.0 & 80.0 & 66.0 & 58.0 & 52.0 & 56.0 & 64.0 & 48.0 & 48.0 & 42.0 & 40.0 & 44.0 & 56.7 \\
Qwen3.6-35B & 54.0 & 44.0 & 42.0 & 40.0 & 14.0 & 4.0 & 2.0 & 10.0 & 14.0 & 10.0 & 6.0 & 0.0 & 20.0 \\
\avg & 70.5 & 71.5 & 62.5 & 57.5 & 24.5 & 34.5 & 45.0 & 38.0 & 30.0 & 32.0 & 20.5 & 20.0 & 42.2 \\
\addlinespace[3pt]
\cpsat \\
Gemini-3-Flash & 98.0 & 100.0 & 100.0 & 100.0 & 100.0 & 98.0 & 98.0 & 100.0 & 96.0 & 100.0 & 96.0 & 96.0 & 98.5 \\
GPT-5-mini & 98.0 & 100.0 & 100.0 & 100.0 & 100.0 & 98.0 & 98.0 & 100.0 & 96.0 & 98.0 & 88.0 & 86.0 & 96.8 \\
DeepSeek-V4-Flash & 98.0 & 100.0 & 100.0 & 100.0 & 100.0 & 98.0 & 98.0 & 100.0 & 96.0 & 100.0 & 96.0 & 100.0 & 98.8 \\
Qwen3.6-35B & 98.0 & 100.0 & 100.0 & 100.0 & 100.0 & 98.0 & 98.0 & 100.0 & 96.0 & 100.0 & 96.0 & 100.0 & 98.8 \\
\avg & 98.0 & 100.0 & 100.0 & 100.0 & 100.0 & 98.0 & 98.0 & 100.0 & 96.0 & 99.5 & 94.0 & 95.5 & 98.2 \\
\bottomrule
\end{tabular}
\caption{\textbf{Per-model results on AsyncPlan-XXL by number of actions, 50 tasks per model and size.} \textbf{(a)} Plan accuracy: every pipeline must satisfy gold action coverage, durations, precedence and optimal makespan; Planner schedules are derived from predicted graphs. \textbf{(b)} Objective-only makespan accuracy: Planner uses deterministic scheduling of its predicted graph and formalizers use returned solver schedules; this diagnostic does not check action coverage, duration correctness or precedence feasibility.}
\label{tab:asyncplanxxl_scaling_results}
\end{table*}

% preamble: \usepackage{booktabs, makecell} (plus the xcolor colors and macros from before)
\begin{table*}[t]
\centering
\small
\renewcommand{\arraystretch}{1.15}
\setlength{\tabcolsep}{4pt}
\renewcommand{\cellalign}{rb}  % right-align makecell header cells
\begin{tabular*}{\linewidth}{@{\extracolsep{\fill}}l rrr rrr@{}}
\toprule
& \multicolumn{3}{c}{\textbf{AsyncHow: makespan accuracy}} & \multicolumn{3}{c}{\textbf{Robotouille: simulator success}} \\
\cmidrule(lr){2-4} \cmidrule(lr){5-7}
\textbf{Model}
& \makecell{\textcolor{plannerblue}{\textsc{Planner}}}
& \makecell{\textcolor{pddlorange}{\textsc{PDDL2.1}}\\\textcolor{pddlorange}{\textsc{Formalizer}}}
& \makecell{\textcolor{cpsatgreen}{\textsc{CP-SAT}}\\\textcolor{cpsatgreen}{\textsc{Formalizer}}}
& \makecell{\textcolor{plannerblue}{\textsc{Planner}}}
& \makecell{\textcolor{pddlorange}{\textsc{PDDL2.1}}\\\textcolor{pddlorange}{\textsc{Formalizer}}}
& \makecell{\textcolor{cpsatgreen}{\textsc{CP-SAT}}\\\textcolor{cpsatgreen}{\textsc{Formalizer}}} \\
\midrule
Gemini-3-Flash    & 88.75 & 95.94 & \textbf{96.88} & 0.0 & \textbf{10.0} & 0.0 \\
GPT-5-mini        & 96.56 & 85.00 & \textbf{97.50} & 0.0 & \textbf{10.0} & 0.0 \\
DeepSeek-V4-Flash & 96.56 & 80.00 & \textbf{98.44} & 0.0 & \textbf{10.0} & 0.0 \\
Qwen3.6-35B       & 95.63 & 55.94 & \textbf{97.19} & 0.0 & \textbf{10.0} & 0.0 \\
\bottomrule
\end{tabular*}
\caption{\textbf{Per-model results on AsyncHow and Robotouille.} Columns compare the \planner, the \pddl and the \cpsat. Makespan accuracy (\S\ref{sec:experimental_setup}) on AsyncHow is computed from the \planner's deterministic scheduling of its predicted graph and from the solver schedules returned by the formalizers; it does not check action coverage, duration correctness or precedence feasibility, so it is not comparable to plan accuracy. Robotouille reports simulator success. Bold marks the best method per row and benchmark.}
\label{tab:async_robo_full_results}
\end{table*}

% preamble: \usepackage{booktabs, makecell} (plus the xcolor colors and macros from before)
\begin{table*}[t]
\centering\scriptsize
\renewcommand{\arraystretch}{1.15}
\setlength{\tabcolsep}{4pt}
\renewcommand{\cellalign}{rb}
\begin{tabular}{@{}l rrr @{\hskip 10pt} rrrr@{}}
\toprule
& \multicolumn{3}{c}{\textbf{Offline}} & \multicolumn{4}{c}{\textbf{Online}} \\
\cmidrule(lr){2-4}\cmidrule(lr){5-8}
\textbf{Model}
& \makecell{\textcolor{plannerblue}{\textsc{Planner}}}
& \makecell{\textcolor{pddlorange}{\textsc{PDDL2.1}}\\\textcolor{pddlorange}{\textsc{Formalizer}}}
& \makecell{\textcolor{cpsatgreen}{\textsc{CP-SAT}}\\\textcolor{cpsatgreen}{\textsc{Formalizer}}}
& \makecell{\textcolor{plannerblue}{\textsc{Planner}}}
& \makecell{\textcolor{pddlorange}{\textsc{PDDL2.1}}\\\textcolor{pddlorange}{\textsc{Formalizer}}}
& \makecell{\textcolor{cpsatgreen}{\textsc{CP-SAT}}\\\textcolor{cpsatgreen}{\textsc{Formalizer}}}
& \makecell{\textcolor{repairpurple}{\textsc{State-aware}}\\\textcolor{repairpurple}{\textsc{Repair}}} \\
\midrule
Gemini-3-Flash    & 40.7 & 52.9 & \textbf{100.0} & 15.0 & 1.4 & 71.4 & \textbf{82.9} \\
GPT-5-mini        & 73.6 & 55.7 & \textbf{99.3}  & 33.6 & 0.0 & 25.0 & \textbf{80.7} \\
DeepSeek-V4-Flash & 17.1 & 27.9 & \textbf{97.9}  & 4.3  & 0.0 & 3.6  & \textbf{80.7} \\
Qwen3.6-35B       & 59.3 & 68.6 & \textbf{97.9}  & 27.9 & 0.0 & 66.4 & \textbf{82.1} \\
\midrule
\rowcolor{black!8}
\textit{Average}  & 47.7 & 51.2 & \textbf{98.8}  & 20.2 & 0.4 & 41.6 & \textbf{81.6} \\
\bottomrule
\end{tabular}
\caption{\textbf{Plan accuracy (\%) for Robo Challenge and Online Robo Challenge, 140 tasks per model.} Columns compare the \planner, the \pddl and the \cpsat; \staterepair is the CP-SAT repair variant of \S\ref{sec:rq3}. Offline evaluation checks durations, grounded execution and rewards. Online evaluation additionally checks true event arrivals, temporal lags and protocol completion; optimization episodes require prefix-conditioned reward optimality. Bold marks the best method per row within each setting.}
\label{tab:robo_challenge_combined}
\end{table*}

% preamble: \usepackage{makecell}  (plus the xcolor colors and macros from before)

\begin{table}[t]
\centering\scriptsize
\setlength{\tabcolsep}{3pt}
\begin{tabular}{lrrr}
\toprule
\textbf{Split}
& \makecell[r]{\textcolor{plannerblue}{\textsc{Planner}}}
& \makecell[r]{\textcolor{pddlorange}{\textsc{PDDL2.1}}\\\textcolor{pddlorange}{\textsc{Formalizer}}}
& \makecell[r]{\textcolor{cpsatgreen}{\textsc{CP-SAT}}\\\textcolor{cpsatgreen}{\textsc{Formalizer}}} \\
\midrule
\multicolumn{4}{l}{\textit{Robo Challenge (offline)}} \\
Easy & 100.0 & 98.8 & 100.0 \\
Hard High-Speedup & 26.2 & 38.8 & 96.2 \\
Hard Multi-Agent & 27.5 & 6.2 & 97.5 \\
Hard Optimization & 3.8 & 0.0 & 100.0 \\
Hard Station & 41.2 & 68.8 & 97.5 \\
Hard Temporal & 58.8 & 48.8 & 100.0 \\
Medium & 76.2 & 97.5 & 100.0 \\
\avg & 47.7 & 51.2 & 98.8 \\
\midrule
\multicolumn{4}{l}{\textit{Online Robo Challenge}} \\
Easy & 66.2 & 0.0 & 70.0 \\
Hard High-Speedup & 8.8 & 0.0 & 55.0 \\
Hard Multi-Agent & 1.2 & 0.0 & 56.2 \\
Hard Optimization & 0.0 & 0.0 & 0.0 \\
Hard Station & 18.8 & 1.2 & 52.5 \\
Hard Temporal & 8.8 & 1.2 & 0.0 \\
Medium & 37.5 & 0.0 & 57.5 \\
\avg & 20.2 & 0.4 & 41.6 \\
\bottomrule
\end{tabular}
\caption{\textbf{Plan accuracy (\%) by split.} Columns are the \planner, the \pddl, and the \cpsat. Each row averages four models on 20 distinct tasks. Online results use true event arrivals and protocol completion. Table~\ref{tab:robo_challenge_combined} reports per-model values.}
\label{tab:robo_challenge_per_split}
\end{table}

% \begin{table}[t]
% \centering
% \small
% \begin{tabular*}{\linewidth}{@{\extracolsep{\fill}}llr@{}}
% \toprule
% \textbf{Method} & \textbf{Model} & \textbf{Acc.} \\
% \midrule
% \multirow{4}{*}{State-Aware CP-SAT Repair}
% & DeepSeek-V4-Flash & 83.6 \\
% & GPT-5-mini & 83.6 \\
% & Gemini-3-flash & 85.7 \\
% & Qwen3.6 35B A3B & 85.0 \\
% \bottomrule
% \end{tabular*}
% \caption{State-aware CP-SAT repair results on Online Robo Challenge explicit tasks. Accuracy is reported as a percentage over 140 tasks per model.}
% \label{tab:online-state-aware-results}
% \end{table}

\subsection{Controlled IR Analysis: Per-Model Breakdown}
\label{app:common_ir_control}

Table~\ref{tab:common_ir_control_full} reports the controlled results of \S\ref{sec:rq2} by model and size. An IR lists only a task's actions, durations, and precedence edges (Figure~\ref{fig:ir_example}). The controls differ in who produces the solver input, which isolates different failure sources:
\begin{itemize}
    \item \textbf{Direct:} the LLM writes the formal specification end to end.
    \item \textbf{NL IR:} the LLM writes a solver-neutral task graph, which is compiled deterministically.
    \item \textbf{Gold IR:} no LLM is involved; the gold task graph and durations are compiled deterministically.
\end{itemize}

% preamble: \usepackage{booktabs, makecell} + \usepackage[table]{xcolor} + the macros from before
\begin{table*}[t]
\centering\scriptsize
\renewcommand{\arraystretch}{1.12}
\setlength{\tabcolsep}{3pt}
\renewcommand{\cellalign}{rb}
\begin{tabular}{@{}l l c @{\hskip 8pt} rrrr @{\hskip 8pt} rrrrr @{\hskip 8pt} r@{}}
\toprule
& & & \multicolumn{4}{c}{\textbf{By model}} & \multicolumn{5}{c}{\textbf{By number of actions}} & \\
\cmidrule(lr){4-7}\cmidrule(lr){8-12}
\textbf{Pipeline} & \textbf{Solver} & \textbf{Retries}
& \makecell{Gemini-\\3-Flash} & \makecell{GPT-5-\\mini} & \makecell{DeepSeek-\\V4-Flash} & \makecell{Qwen3.6-\\35B}
& 10 & 30 & 50 & 70 & 100 & \textbf{Avg.} \\
\midrule
\multicolumn{13}{@{}l}{\textit{The model writes the formal specification (Direct)}} \\
\textcolor{pddlorange}{\textsc{PDDL2.1}} & OPTIC & 0 & 19.2 & 0.4 & 1.6 & 0.4 & 26.0 & 1.0 & 0.0 & 0.0 & 0.0 & 5.4 \\
\textcolor{cpsatgreen}{\textsc{CP-SAT}} & CP-SAT & 0 & 94.0 & 80.8 & 48.0 & 94.0 & 99.0 & 98.0 & 80.5 & 64.0 & 54.5 & 79.2 \\
\midrule
\multicolumn{13}{@{}l}{\textit{The model writes the task graph, which is compiled deterministically (NL IR)}} \\
\textcolor{pddlorange}{\textsc{PDDL2.1}} & OPTIC & 0 & 94.0 & 82.0 & 82.0 & 68.4 & 83.5 & 92.0 & 92.0 & 84.5 & 56.0 & 81.6 \\
\textcolor{cpsatgreen}{\textsc{CP-SAT}} & CP-SAT & 0 & 93.2 & 82.4 & 82.0 & 68.4 & 83.5 & 92.0 & 92.0 & 84.0 & 56.0 & 81.5 \\
\textcolor{plannerblue}{\textsc{Planner}} & longest path & 0 & 90.0 & 66.0 & 75.6 & 83.2 & 94.0 & 97.0 & 84.0 & 74.0 & 44.5 & 78.7 \\
\midrule
\multicolumn{13}{@{}l}{\textit{The task graph is given and compiled deterministically; no model in the loop (Gold IR)}} \\
\textcolor{pddlorange}{\textsc{PDDL2.1}} & OPTIC & -- & -- & -- & -- & -- & 100.0 & 100.0 & 100.0 & 100.0 & 100.0 & 100.0 \\
\textcolor{pddlorange}{\textsc{PDDL2.1}} & POPF & -- & -- & -- & -- & -- & 100.0 & 100.0 & 100.0 & 100.0 & 100.0 & 100.0 \\
\textcolor{cpsatgreen}{\textsc{CP-SAT}} & CP-SAT & -- & -- & -- & -- & -- & 100.0 & 100.0 & 100.0 & 100.0 & 100.0 & 100.0 \\
\midrule
\multicolumn{13}{@{}l}{\textit{Direct, main experimental setting}} \\
\textcolor{pddlorange}{\textsc{PDDL2.1}} & OPTIC & 3 & 38.4 & 22.4 & 43.2 & 6.0 & 57.0 & 20.0 & 32.5 & 19.0 & 9.0 & 27.5 \\
\textcolor{cpsatgreen}{\textsc{CP-SAT}} & CP-SAT & 3 & 96.8 & 84.0 & 97.6 & 97.2 & 99.5 & 99.5 & 97.5 & 90.5 & 82.5 & 93.9 \\
\bottomrule
\end{tabular}
\caption{\textbf{Plan accuracy (\%) of the IR controls by model and by size.} Results are on the five AsyncPlan-XXL sizes shared by all controls. In Direct, the model writes the formal specification itself; in NL IR, it writes only the task graph and a deterministic compiler produces the PDDL2.1 or CP-SAT code; in Gold IR, the task graph is given and no model is involved. Zero-retry Direct and NL IR \Formalizer rows use the IR-control configuration (\texttt{max\_tokens}=8192); the bottom Direct rows use the main configuration (\texttt{max\_tokens}=32768, up to three feedback retries).}
\label{tab:common_ir_control_full}
\end{table*}

\paragraph{Comparison with the main results.}
Table~\ref{tab:common_ir_control_full} uses the same models and the $N \in \{10,30,50,70,100\}$ subset of Table~\ref{tab:asyncplanxxl_scaling_results}, and its three-retry Direct rows reproduce those averages. The zero-retry Direct and NL IR \Formalizer rows come from the IR-control run, whose launch configuration sets the per-call output-token limit to \texttt{max\_tokens}=8192. The bottom Direct rows use the main AsyncPlan-XXL \Formalizer configuration, with \texttt{max\_tokens}=32768 and up to three feedback retries. These token values describe the launch-script settings; the archived results do not record the effective provider-side limits. Direct CP-SAT reaches 79.2\% in the control run and 93.9\% in the main setting; Direct PDDL2.1 reaches 5.4\% and 27.5\%, respectively. These comparisons change both the configured token budget and the retry allowance, so they do not isolate the effect of either. Within the zero-retry control run shown in Figure~\ref{fig:ir_decomposition}, an intermediate task graph helps CP-SAT for some models and hurts it for others: DeepSeek-V4-Flash improves from 48.0\% to 82.0\%, while Qwen3.6-35B drops from 94.0\% to 68.4\%. For PDDL2.1, compiling model-extracted IR into OPTIC reaches 81.6\% on average, far above the 5.4\% of Direct.

\paragraph{IR compilation, validation, and the Gold IR ceiling.}
\label{app:ir_compilation}
Both compilers receive the same actions, durations, and precedence edges.
% The PDDL compiler adds an explicit \texttt{total-time} minimization objective, keeps the original duration units, and applies no schedule retiming.
All 953 acyclic NL IRs yield schedules that are valid and optimal for their input graph, so remaining errors stem from graph extraction, not compilation or solving. Against the gold task, 81.6\% of the 1,000 NL IR records succeed; failures, in order of assignment, are missing IR (4.6\%), cycles (0.1\%), duration mismatches (6.2\%), and precedence violations (7.5\%).

On the same NL IR, OPTIC and CP-SAT reach 81.6\% and 81.5\%. The three records where they differ omit gold dependencies, so a schedule valid for the underspecified graph may be invalid for the gold task; plan accuracy and faithful extraction can thus diverge. Given the gold graph, both solvers reach 100\% on all 250 unique tasks, measuring compilation and solving alone since no model is involved.

\paragraph{Check with an independent planner.}
To test whether this ceiling depends on OPTIC, we run POPF~\citep{coles2010popf} (ROSPlan build, pinned by SHA-256; no model calls or remote services) on the same 250 compiled gold problems with a 10-second wall-clock limit. POPF returns a plan for every problem without timeouts, and re-running the binary reproduces all 250 plans exactly. Acceptance depends on how outputs are checked:
\begin{itemize}
    \item \textbf{Strict makespan equality:} 54.0\% are accepted, since POPF separates action start times by a small $\epsilon$.
    \item \textbf{Start times snapped to the duration grid:} all 250 are valid and makespan-optimal at every graph size.
    \item \textbf{Original seconds encoding:} 86.0\% are accepted; the other 35 fail from a numeric overflow in the plan printer, not from search.
\end{itemize}
This control relies on the fixed-duration, precedence-only structure of these tasks and need not generalize to all PDDL backends, but it shows that the gold-structure ceiling in Table~\ref{tab:common_ir_control_full} is not specific to one planner.

\begin{figure*}[t]
\begin{tcolorbox}[colback=gray!8, colframe=gray!50, title={IR example: \textit{Hiatal hernia} (AsyncPlan-XXL, 10 actions; IR extracted by Gemini-3-Flash)}, fonttitle=\bfseries\small, left=4pt, right=4pt, top=4pt, bottom=4pt]
\small
\textbf{Task (abridged):}
S1. Schedule a GP consultation (10\,min) \quad
S2. Physical examination (30\,min) \quad
S3. Research imaging centers (1\,h) \quad
S4. Verify insurance coverage (20\,min) \quad
S5. Fast before the procedure (8\,h) \\
S6. Upper GI X-ray series (45\,min) \quad
S7. Specialist referral (5\,min) \quad
S8. Travel to the surgical center (1\,h) \quad
S9. Pick up prescription (15\,min) \quad
S10. Consult the gastroenterologist (45\,min)

\smallskip
\textbf{Stated constraints:} S1$\to$S2, \; S2$\to$S7, \; S2$\to$S9, \; S3$\to$S4, \; S5$\to$S6, \; S7$\to$S10, \; S8$\to$S6, \; S10$\to$S8

\smallskip
\textbf{IR written by the model} (action descriptions and the goal list omitted):\\
{\footnotesize\ttfamily
\{"actions": [\{"id": "step1", "duration": 600\}, \{"id": "step2", "duration": 1800\}, \{"id": "step3", "duration": 3600\}, \{"id": "step4", "duration": 1200\}, \{"id": "step5", "duration": 28800\}, \{"id": "step6", "duration": 2700\}, \{"id": "step7", "duration": 300\}, \{"id": "step8", "duration": 3600\}, \{"id": "step9", "duration": 900\}, \{"id": "step10", "duration": 2700\}],\\
\ "precedence": [\{"before": "step1", "after": "step2"\}, \{"before": "step2", "after": "step7"\}, \{"before": "step2", "after": "step9"\}, \{"before": "step3", "after": "step4"\}, \{"before": "step5", "after": "step6"\}, \{"before": "step7", "after": "step10"\}, \{"before": "step8", "after": "step6"\}, \{"before": "step10", "after": "step8"\}],\\
\ "objective": "minimize\_makespan"\}}

\smallskip
\textbf{Compiled PDDL2.1} (one of the ten actions, and the problem's goal and metric):\\
{\footnotesize\ttfamily
(:durative-action step6 :parameters () :duration (= ?duration 2700)\\
\ \ :condition (and (at start (ready)) (at start (done\_step5)) (at start (done\_step8)))\\
\ \ :effect (and (at end (done\_step6))))\\
(:goal (and (done\_step1) \ldots{} (done\_step10)))\quad(:metric minimize (total-time))}

\smallskip
\textbf{Compiled CP-SAT:} each action becomes an interval of its duration, and each precedence edge requires the earlier action to end before the later one starts; the objective minimizes the makespan.

\smallskip
\textbf{Result:} OPTIC and CP-SAT both return the gold optimal makespan of 31{,}500\,s (8\,h\,45\,min), set by the 8-hour fast (S5) followed by the X-ray (S6).
\end{tcolorbox}
\caption{\textbf{An NL IR example.} The model writes only the IR; deterministic compilers produce the PDDL2.1 and CP-SAT inputs. Here the extracted IR matches the gold task graph exactly.}
\label{fig:ir_example}
\end{figure*}

\subsection{Stronger Models and Few-Shot Prompting}

\paragraph{Stronger-model comparison.}
\label{app:strong_model}
Table~\ref{tab:strong_model_scaling} reports GPT-5.6 Sol (API-default temperature) on all 50 tasks at $N \in \{10,30,50,70,100\}$, the same sizes as the four-model baseline. Saved outputs are graded with the main evaluators, including deterministic scheduling and the common gold-constraint validation for Planner; model-reported makespan is not scored. All methods share the same task input and three-retry Formalizer protocol, and Planner receives no solver feedback or model-side tools. Planner and CP-SAT both average 98.0\%, differing by one task only at $N{=}50$ (Planner higher) and $N{=}100$ (CP-SAT higher), while PDDL2.1 reaches 93.2\%.
% The available Planner launch script specifies 16{,}384 output tokens, but the saved strong-model summaries do not record the actual per-run token setting; the four-model baseline also retains the Qwen budget exception.

% preamble: \usepackage{booktabs} + \usepackage[table]{xcolor} + the color definitions from before
\begin{table}[t]
\centering\scriptsize
\renewcommand{\arraystretch}{1.12}
\setlength{\tabcolsep}{3.5pt}
\begin{tabular}{@{}l rrrrr @{\hskip 8pt} r@{}}
\toprule
& \multicolumn{5}{c}{\textbf{Number of actions}} & \\
\cmidrule(lr){2-6}
\textbf{Model} & 10 & 30 & 50 & 70 & 100 & \textbf{Avg.} \\
\midrule
\multicolumn{7}{@{}l}{\textcolor{plannerblue}{\textsc{Planner}}} \\[-1pt]
\rowcolor{black!8}
\hspace{0.6em}\textit{4-model avg.} & 94.0 & 97.0 & 84.0 & 74.0 & 44.5 & 78.7 \\
\hspace{0.6em}GPT-5.6 Sol & 100.0 & 100.0 & 98.0 & 94.0 & 98.0 & 98.0 \\
\addlinespace[3pt]
\multicolumn{7}{@{}l}{\textcolor{pddlorange}{\textsc{PDDL2.1}}} \\[-1pt]
\rowcolor{black!8}
\hspace{0.6em}\textit{4-model avg.} & 57.0 & 20.0 & 32.5 & 19.0 & 9.0 & 27.5 \\
\hspace{0.6em}GPT-5.6 Sol & 98.0 & 90.0 & 96.0 & 92.0 & 90.0 & 93.2 \\
\addlinespace[3pt]
\multicolumn{7}{@{}l}{\textcolor{cpsatgreen}{\textsc{CP-SAT}}} \\[-1pt]
\rowcolor{black!8}
\hspace{0.6em}\textit{4-model avg.} & 99.5 & 99.5 & 97.5 & 90.5 & 82.5 & 93.9 \\
\hspace{0.6em}GPT-5.6 Sol & 100.0 & 100.0 & 96.0 & 94.0 & 100.0 & 98.0 \\
\bottomrule
\end{tabular}
\caption{\textbf{Stronger-model comparison across pipelines on AsyncPlan-XXL, plan accuracy (\%).} Each size uses 50 tasks. Grey rows give the four-model average from Table~\ref{tab:asyncplanxxl_scaling_results}, computed over the same five sizes. Model-reported makespan is excluded. API settings are described in the text.}
\label{tab:strong_model_scaling}
\end{table}

\paragraph{Few-shot prompting for the PDDL2.1 Formalizer.}
\label{app:fewshot}
Table~\ref{tab:fewshot_pddl} reports the few-shot experiment, which varies PDDL exemplars on a separate set of zero-shot runs (50 tasks per cell) rather than the run selection of the four-model main table. Plan accuracy is 33.6\% zero-shot, 44.4\% one-shot, and 42.7\% two-shot. Three one-shot records with unresolved action mappings count as failures; counting them as successes gives 45.1\%. One-shot thus improves over zero-shot, but two-shot adds nothing further, indicating sensitivity to exemplars rather than monotonic improvement or robustness across arbitrary prompts.

% % Generated by scripts/build_pddl_replay_paper.py; all scores are percentages.
% \begin{table}[t]
% \centering\scriptsize
% \setlength{\tabcolsep}{3pt}
% \begin{tabular}{lrrrrrr}
% \toprule
% Model & 10 & 30 & 50 & 70 & 100 & Avg. \\
% \midrule
% Gemini 3 Flash & 94.0 & 2.0 & 0.0 & 0.0 & 0.0 & 19.2 \\
% GPT-5-mini & 0.0 & 2.0 & 0.0 & 0.0 & 0.0 & 0.4 \\
% DeepSeek-V4-Flash & 8.0 & 0.0 & 0.0 & 0.0 & 0.0 & 1.6 \\
% Qwen3.6 35B A3B & 2.0 & 0.0 & 0.0 & 0.0 & 0.0 & 0.4 \\
% \bottomrule
% \end{tabular}
% \caption{Direct PDDL2.1 plan accuracy by model and task size, 50 tasks per cell. Averages cover these five sizes.}
% \label{tab:pddl_zero_retry}
% \end{table}

% preamble: \usepackage{booktabs} + \usepackage[table]{xcolor} + the macros from before
\begin{table}[t]
\centering\scriptsize
\renewcommand{\arraystretch}{1.15}
\setlength{\tabcolsep}{2.4pt}
\begin{tabular}{@{}r rrr @{\hskip 5pt} rrr @{\hskip 5pt} rrr@{}}
\toprule
& \multicolumn{3}{c}{\textbf{Gemini-3-Flash}} & \multicolumn{3}{c}{\textbf{GPT-5-mini}} & \multicolumn{3}{c}{\textbf{DeepSeek-V4-Flash}} \\
\cmidrule(lr){2-4}\cmidrule(lr){5-7}\cmidrule(lr){8-10}
\diagbox[font=\scriptsize, innerleftsep=1pt, innerrightsep=1pt, height=1.8em]{\#Act.}{\#Shot} & 0 & 1 & 2 & 0 & 1 & 2 & 0 & 1 & 2 \\
\midrule
10  & \textbf{76.0} & 66.0 & 62.0 & 56.0 & \textbf{68.0} & 66.0 & 74.0 & \textbf{98.0} & 90.0 \\
50  & 6.0 & 44.0 & \textbf{64.0} & 8.0 & \textbf{10.0} & 2.0 & 52.0 & \textbf{62.0} & 56.0 \\
100 & 0.0 & 20.0 & \textbf{24.0} & 0.0 & 0.0 & 0.0 & 30.0 & \textbf{32.0} & 20.0 \\
\midrule
\rowcolor{black!8}
\textit{Avg.} & 27.3 & 43.3 & \textbf{50.0} & 21.3 & \textbf{26.0} & 22.7 & 52.0 & \textbf{64.0} & 55.3 \\
\bottomrule
\end{tabular}
\caption{\textbf{Plan accuracy (\%) for few-shot \pddl, 50 tasks per cell.} Rows give the number of actions and columns the number of exemplars (0, 1, 2). Exemplars come from AsyncHow with exact-text overlap checks. Bold marks the best number of shots per row within each model. Averaged over all three models, accuracy is 33.6 (0-shot), 44.4 (1-shot) and 42.7 (2-shot).}
\label{tab:fewshot_pddl}
\end{table}

\FloatBarrier
\section{Evaluation Protocol}
\label{app:eval_protocol}

\subsection{Evaluation Details}
\label{app:evaluation_details}

Section~\ref{sec:benchmarks} defines plan accuracy; we add only the implementation details it omits. All three pipelines share one schedule validator, with duration and precedence tolerances of $10^{-6}$ seconds and an objective tolerance of $0.5$ seconds to allow for OPTIC's millisecond event separation. On AsyncPlan-XXL, Planner's earliest-start schedule is computed solely from its predicted graph and the gold constraints are used only for validation; exact edge matching is not required, and graph equivalence (durations and transitive closure) is recorded separately in the per-instance audit. Model-reported makespan is an arithmetic diagnostic and does not determine plan accuracy.

On Robo Challenge, durations must match within $10^{-6}$ seconds, and optimization tasks accept any completed candidate set that attains the gold reward within the inventory and deadline limits, rather than the witness's particular set. Offline evaluation uses the 1,680 model--task records whose inputs match the benchmark rendering. Online success additionally requires completing the generation protocol, so a failed generation or solver stage is a failure even if its surviving execution prefix is feasible; online makespan optimality is a separate diagnostic.

Faithfulness diagnostics use transitive closures for precedence precision and recall, so redundant direct edges are harmless. PDDL2.1 diagnostics parse the supported positive, conjunctive durative-action subset and recover action identities from saved mappings, explicit bindings, or documented semantic review; ambiguous or unsupported cases score zero, which marks them as unverified rather than shown to be incorrect.

\subsection{Faithfulness diagnostics}
\label{app:error_analysis}

\begin{table*}[t]
\centering\small
\setlength{\tabcolsep}{5pt}
\begin{tabular}{lrrr}
\toprule
& \makecell{\textcolor{plannerblue}{\textsc{Planner}}}
& \makecell{\textcolor{pddlorange}{\textsc{PDDL2.1}}\\\textcolor{pddlorange}{\textsc{Formalizer}}}
& \makecell{\textcolor{cpsatgreen}{\textsc{CP-SAT}}\\\textcolor{cpsatgreen}{\textsc{Formalizer}}} \\
\midrule
\multicolumn{4}{@{}l}{\textit{Each diagnostic (mean score)}} \\
Action count & 90.5 & 65.3 & 99.9 \\
Per-action durations & 90.3 & 64.6 & 99.7 \\
Goal-action coverage & 90.5 & 34.1 & 99.9 \\
Precedence precision & 90.4 & 58.3 & 99.1 \\
Precedence recall & 90.1 & 60.2 & 99.1 \\
\midrule
\multicolumn{4}{@{}l}{\textit{Whole records}} \\
All five verified & 79.1 & 16.7 & 93.9 \\
Plan accuracy, among verified records & 100.0 & 100.0 & 100.0 \\
Plan accuracy, all records & 81.7 & 30.8 & 94.0 \\
\bottomrule
\end{tabular}
\caption{\textbf{Faithfulness diagnostics on the AsyncPlan-XXL main experiment, in \% of the 2,400 records per pipeline.} Each diagnostic row is the mean score over all records, with outputs whose structure cannot be read back scored as zero. A record is verified when all five diagnostics score 1. Unreadable outputs are unverified, not shown to be wrong.}
\label{tab:faithfulness}
\end{table*}

% preamble: \usepackage{booktabs}  (plus the xcolor colors from before)
\begin{table}[t]
\centering\scriptsize
\renewcommand{\arraystretch}{1.1}
\setlength{\tabcolsep}{3pt}
\begin{tabular}{@{}r rrr @{\hskip 8pt} rrr@{}}
\toprule
& \multicolumn{3}{c}{\textbf{Precedence Recall}} & \multicolumn{3}{c}{\textbf{Objective-only Acc.}} \\
\cmidrule(lr){2-4}\cmidrule(lr){5-7}
\textbf{Actions}
& \textcolor{plannerblue}{\textsc{Planner}} & \textcolor{pddlorange}{\textsc{PDDL}} & \textcolor{cpsatgreen}{\textsc{CP}}
& \textcolor{plannerblue}{\textsc{Planner}} & \textcolor{pddlorange}{\textsc{PDDL}} & \textcolor{cpsatgreen}{\textsc{CP}} \\
\midrule
5   & \textbf{100.0} & 77.4 & 95.0           & 97.5 & 70.5 & \textbf{98.0} \\
10  & 94.4           & 79.0 & \textbf{99.5}  & 94.5 & 71.5 & \textbf{100.0} \\
15  & \textbf{99.5}  & 72.9 & 98.5           & 99.5 & 62.5 & \textbf{100.0} \\
20  & \textbf{99.5}  & 74.3 & \textbf{99.5}  & 98.5 & 57.5 & \textbf{100.0} \\
30  & 98.0           & 64.5 & \textbf{99.5}  & 97.5 & 24.5 & \textbf{100.0} \\
40  & 95.8           & 60.6 & \textbf{99.0}  & 93.5 & 34.5 & \textbf{98.0} \\
50  & 92.6           & 61.5 & \textbf{99.5}  & 89.0 & 45.0 & \textbf{98.0} \\
60  & 92.7           & 48.3 & \textbf{100.0} & 92.0 & 38.0 & \textbf{100.0} \\
70  & 89.8           & 50.3 & \textbf{99.5}  & 83.5 & 30.0 & \textbf{96.0} \\
80  & 85.4           & 46.5 & \textbf{100.0} & 81.0 & 32.0 & \textbf{99.5} \\
90  & 66.9           & 45.7 & \textbf{99.5}  & 58.5 & 20.5 & \textbf{94.0} \\
100 & 66.9           & 41.1 & \textbf{99.5}  & 57.5 & 20.0 & \textbf{95.5} \\
\bottomrule
\end{tabular}
\caption{\textbf{Precedence recall and objective-only accuracy (\%) by size.} Recall is the mean precedence recall over reachable pairs, across all 200 records per size; outputs whose structure cannot be read back score zero. Objective-only columns keep all 200 records per size and do not imply valid schedules. Bold marks the best method per row and metric.}
\label{tab:stepwise-diagnostics}
\end{table}

Table~\ref{tab:faithfulness} reports the five faithfulness diagnostics of \S\ref{sec:benchmarks} on the AsyncPlan-XXL main experiment, first for each diagnostic and then for whole records. Each diagnostic has a score between 0 and 1, and a record passes a diagnostic when its score is 1. Computing a diagnostic requires reading the task structure back from the output; for PDDL2.1 this requires parsing the supported durative-action subset and recovering action identities, as Appendix~\ref{app:evaluation_details} describes. When the structure cannot be read back, the diagnostic scores zero, so scores average over all 2,400 records of a pipeline, the three pipelines share one denominator, and each score is a lower bound: an unreadable output is unverified, not shown to be wrong.

At the record level, a record is verified when it passes all five diagnostics. In our data every verified record also has a correct plan, but verification is not necessary for one, because a record may be unreadable, or an extra precedence edge may lower precision without changing the optimal schedule.

Table~\ref{tab:stepwise-diagnostics} reports the precedence-recall diagnostic by size alongside objective-only accuracy. Unsupported interpretations are not assigned a semantic error category.

Error analysis distinguishes observable schedule violations from interpretation-dependent faithfulness diagnostics. Tables~\ref{tab:faithfulness} and~\ref{tab:stepwise-diagnostics} report the supported diagnostic coverage. Qualitative examples below illustrate possible failures; they are not estimates of their prevalence. Online validation checks actual arrivals, durations, lags and protocol completion independently of the runner\textquotesingle s success flag.
Concrete illustrative failure examples are provided in \S\ref{app:qualitative_errors}.

\subsection{Uncertainty Estimates}
\label{app:uncertainty}

\begin{table}[t]
\centering\scriptsize
\renewcommand{\arraystretch}{1.12}
\setlength{\tabcolsep}{4pt}
\begin{tabular}{@{}l c @{\hskip 6pt} c @{\hskip 6pt} r@{}}
\toprule
\textbf{Pipeline} & \textbf{Plan Accuracy} & \textbf{95\% CI} & \textbf{Standard Error} \\
\midrule
\multicolumn{4}{@{}l}{\textit{AsyncPlan-XXL, 600 tasks}} \\[-1pt]
\hspace{0.6em}\textcolor{plannerblue}{\textsc{Planner}}  & 81.7 & [80.2, 83.0] & 0.71 \\
\hspace{0.6em}\textcolor{pddlorange}{\textsc{PDDL2.1}}   & 30.8 & [29.2, 32.4] & 0.84 \\
\hspace{0.6em}\textcolor{cpsatgreen}{\textsc{CP-SAT}}    & 94.0 & [92.8, 95.1] & 0.60 \\
\midrule
\multicolumn{4}{@{}l}{\textit{Robo Challenge, 140 tasks}} \\[-1pt]
\hspace{0.6em}\textcolor{plannerblue}{\textsc{Planner}}  & 47.7 & [45.5, 49.8] & 1.09 \\
\hspace{0.6em}\textcolor{pddlorange}{\textsc{PDDL2.1}}   & 51.2 & [49.3, 53.2] & 1.00 \\
\hspace{0.6em}\textcolor{cpsatgreen}{\textsc{CP-SAT}}    & 98.8 & [97.9, 99.5] & 0.44 \\
\midrule
\multicolumn{4}{@{}l}{\textit{Online Robo Challenge, 140 tasks}} \\[-1pt]
\hspace{0.6em}\textcolor{plannerblue}{\textsc{Planner}}  & 20.2 & [17.9, 22.7] & 1.23 \\
\hspace{0.6em}\textcolor{pddlorange}{\textsc{PDDL2.1}}   & 0.4  & [0.0, 0.9]   & 0.25 \\
\hspace{0.6em}\textcolor{cpsatgreen}{\textsc{CP-SAT}}    & 41.6 & [39.6, 43.6] & 0.97 \\
\hspace{0.6em}\textcolor{repairpurple}{\textsc{State-aware repair}} & 81.6 & [78.8, 84.1] & 1.36 \\
\bottomrule
\end{tabular}
\caption{\textbf{Task-cluster bootstrap uncertainty conditional on the fixed models and configurations.} The \pddl and the \cpsat are shown as PDDL2.1 and CP-SAT, and state-aware repair is the CP-SAT repair variant of \S\ref{sec:rq3}. Pipelines and models stay paired on each sampled task; resampling is stratified by size or split. Accuracy is in percent; CI and SE are in percentage points.}
\label{tab:uncertainty}
\end{table}

We compute the main aggregates from the saved per-instance outputs, using the same run selection and evaluators as the results tables. We draw 10{,}000 bootstrap replicates (seed 42), resampling task IDs within each size or split and retaining all models and pipelines together for each sampled task. The four models are fixed; they are not treated as independent samples of model capability. AsyncPlan-XXL contains 600 task clusters (50 per size); each grounded benchmark contains 140 (20 per split). Sizes and splits receive equal weight. Table~\ref{tab:uncertainty} reports percentile 95\% intervals and bootstrap standard errors.

These intervals quantify task-sampling uncertainty conditional on the tested models, prompts, budgets, and validators. They do not cover prompt variation, repeated model generations, solver randomness, or systematic evaluation errors. On Robo Challenge, the PDDL2.1--Planner gap is 3.5 points, with a paired interval of [0.9, 6.2]; the CP-SAT advantage is larger. In the stronger-model comparison, the CP-SAT--Planner gap is 0.0 points with a paired interval of [$-$1.2, 1.2], giving no evidence of an average difference under that configuration.

\paragraph{Split-level comparisons.}
Each split has only 20 distinct tasks; four models do not increase this to 80 independent tasks. Applying the same task-cluster bootstrap within Robo splits gives PDDL2.1 intervals of [63.7, 73.8] on \textit{Station}, [31.2, 46.2] on \textit{Speedup}, [1.2, 11.2] on \textit{Multi-Agent}. These descriptive contrasts do not justify treating every smaller split difference as reliable; no correction for multiple comparisons is applied.

\subsection{Robotouille Failure Modes}
\label{app:robotouille}

Table~\ref{tab:async_robo_full_results} gives the per-model results: the same 10 base-layout tasks with each of four models, i.e., 40 repeated evaluations per pipeline rather than 40 independent tasks. Success means the emitted action sequence executes in the Robotouille simulator and satisfies the goals; optimal makespan is not required, and missing plans and solver or simulator failures stay in the denominator. PDDL2.1 succeeds on 10.0\% of tasks for every model, and Planner and CP-SAT on 0.0\%. The PDDL2.1 pipeline runs without Robotouille-specific repairs or a fallback solver, and plans pass through the temporal-plan-to-action-sequence adapter. The failures below are qualitative cases, not estimates of prevalence.

\paragraph{Robot-state and object-location preconditions.} CP-SAT and Planner outputs abstract away the robot's physical state, so the first step already fails, and a sequential version of the same task would fail identically; this is not an asynchronous-planning failure. PDDL2.1 tries to encode these preconditions but typically assigns wrong stations.

\begin{tcolorbox}[colback=gray!8, colframe=gray!50, title=Simulator errors, fonttitle=\bfseries\small, left=4pt, right=4pt, top=4pt, bottom=4pt]
\small
\textbf{Hand state} (step 0): \texttt{precondition (nothing, robot\_1) not satisfied}. The plan starts with \texttt{pick-up} but never asserts an empty hand.

\smallskip
\textbf{Location} (step 0--1): \texttt{precondition (loc, robot\_1, table\_1) not satisfied}. An action at \texttt{table\_1} is issued without a preceding \texttt{move}.

\smallskip
\textbf{Object location}: \texttt{pick-up-container robot\_1 pot\_1 table\_2: (container\_at pot1 table2)} is unmet because the pot was never moved there.
\end{tcolorbox}

\paragraph{Missing repetitions.} Cooking, cutting and frying must each be repeated a fixed number of times (e.g., three \texttt{cook} calls). All three pipelines often emit one, so a later \texttt{stack} fails on \texttt{(is\_cooked chicken\_1)}.

\paragraph{Why Robo Challenge abstracts these away.} The failures are consistent across all four LLMs and three pipelines: the bottleneck is tracking robot state, not ordering, parallelism or resource coordination. Robo Challenge replaces robot movement with station capacity and resource constraints, so the simulator checks resource feasibility without object or robot positions and the comparison isolates asynchronous planning.

\FloatBarrier
\section{Qualitative Error Examples}
\label{app:qualitative_errors}

\begin{figure}[t]
\centering
\includegraphics[width=\columnwidth]{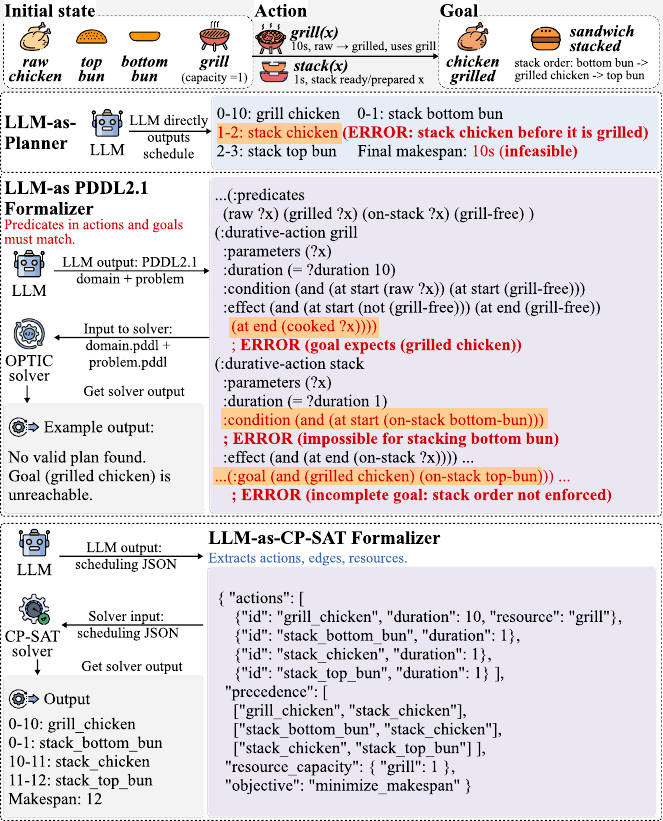}
\caption{\textbf{Illustrative outputs on one state-constrained cooking task.} \planner schedules stacking before grilling finishes; the PDDL2.1 example has inconsistent predicates and unmet preconditions; the CP-SAT example exposes actions, durations, precedence and capacity. These illustrate pipelines, not failure frequency.}
\label{fig:method-error-overview}
\end{figure}

Examples 1--5 illustrate action-effect mismatch, an error in model-reported makespan, duration conversion, circular dependencies, and incomplete goals or dependencies. Example~2 concerns the model-reported arithmetic diagnostic; it is not a failure under the Planner metric when the extracted graph is otherwise correct.
Examples 6--7 illustrate execution and specification errors in the grounded settings; Figure~\ref{fig:method-error-overview} summarizes the three pipelines on one cooking task.

\paragraph{Example 1: PDDL2.1 Formalizer adds spurious preconditions (action-effect mismatch).}
Consider the following 5-step cooking task from AsyncPlan-XXL ($N{=}5$).

\begin{tcolorbox}[colback=gray!8, colframe=gray!50, title=Task: \textit{Eggless Chocolate Cake}, fonttitle=\bfseries\small, left=4pt, right=4pt, top=4pt, bottom=4pt]
\small
\textbf{Steps and durations:}
S1. Whisk flour, cocoa, sugar \quad (5 min) \quad
S2. Preheat oven \quad (15 min) \quad
S3. Sift dry ingredients \quad (3 min) \quad
S4. Grease pans \quad (2 min) \quad
S5. Mix wet ingredients and bake \quad (45 min)

\smallskip
\textbf{Ordering constraints:} S3 $\to$ S1, \; S1 $\to$ S5, \; S2 $\to$ S4

\smallskip
\textbf{Gold optimal makespan:} S3(3\,min) $\to$ S1(5\,min) $\to$ S5(45\,min) $=$ \textbf{53\,min}.\\
S2 and S4 run in parallel along the critical path.
\end{tcolorbox}

\noindent The correct formalization encodes exactly three precedence edges: S3$\to$S1, S1$\to$S5, S2$\to$S4.
CP-SAT Formalizer (GPT-5-mini) does this correctly and returns the optimal 53-minute schedule.
PDDL2.1 Formalizer (same model) encodes step S5 as follows:

\begin{tcolorbox}[colback=gray!8, colframe=gray!50, title=PDDL2.1 Formalizer output (step S5), fonttitle=\bfseries\small, left=4pt, right=4pt, top=4pt, bottom=4pt]
\begin{alltt}\scriptsize
(:durative-action do\_step5\_mix\_and\_bake
  :parameters ()
  :duration (= ?duration 2700)
  :condition (and (at start (step\_pending step5))
                  (at start (s1\_done))
                  (at start (s2\_done))  \textit{; <- spurious}
                  (at start (s3\_done))  \textit{; implied via S1}
                  (at start (s4\_done))) \textit{; <- spurious}
  :effect (and (at start (not (step\_pending step5)))
               (at end (step\_done step5))
               (at end (s5\_done))))
\end{alltt}
\end{tcolorbox}

\noindent The action requires \texttt{s2\_done} and \texttt{s4\_done} in addition to the correct \texttt{s1\_done}, effectively adding the false constraints S2$\to$S5 and S4$\to$S5.
OPTIC now serializes the S2$\to$S4 chain before S5 can start, yielding a predicted makespan of 62 minutes instead of 53 minutes.

\paragraph{Example 2: Planner miscalculates makespan at AND-join nodes.}

Even when Planner correctly extracts every stated ordering constraint, it can still compute the wrong makespan because it fails to account for AND-join semantics, where a node can start only after \emph{all} of its predecessors have completed.

\begin{tcolorbox}[colback=gray!8, colframe=gray!50, title=Task: \textit{Breakfast in Bed} (Gemini-3-Flash on AsyncHow), fonttitle=\bfseries\small, left=4pt, right=4pt, top=4pt, bottom=4pt]
\small
\textbf{Steps:}
S1. Heat up pan (5\,min) \quad
S2. Crack eggs (3\,min) \quad
S3. Whisk eggs (3\,min) \\
S4. Pour whisked eggs (10\,s) \quad
S5. Scramble while cooking (10\,min) \quad
S6. Add butter (10\,s) \quad
S7. Plate eggs (10\,s)

\smallskip
\textbf{Stated constraints:} S1$\to$S6, \; S2$\to$S3, \; S3$\to$S4, \; S4$\to$S5, \; S5$\to$S7, \; S6$\to$S4

\smallskip
\textbf{Gold makespan:} 980\,s \quad \textbf{Predicted makespan:} 930\,s
\end{tcolorbox}

\noindent The model's output JSON contains all six correct dependencies.
Step S4 is an AND-join: it requires \emph{both} the chain S1$\to$S6$\to$S4 (310\,s) \emph{and} the chain S2$\to$S3$\to$S4 (360\,s) to finish.
Step S4 therefore cannot start until $t = 360$\,s, giving a gold makespan of $360 + 10 + 600 + 10 = 980$\,s.
The model-reported 930\,s is consistent with following S1$\to$S6$\to$S4$\to$S5$\to$S7 while overlooking the later completion of S2$\to$S3. This observation does not reveal the model's internal calculation.
Deterministic longest-path computation on the correct graph returns 980\,s, so this self-reported makespan error does not count against Planner plan accuracy. It is also distinct from extracting an incorrect action duration.

\paragraph{Example 3: Planner misreads a time unit (wrong duration).}

A single time-unit confusion causes an error that dominates the makespan even when every dependency is correct.

\begin{tcolorbox}[colback=gray!8, colframe=gray!50, title=Task: \textit{Writer's Dream} (GPT-5-mini on AsyncHow), fonttitle=\bfseries\small, left=4pt, right=4pt, top=4pt, bottom=4pt]
\small
\textbf{Steps:}
S1. Search the Internet for popular genres (20\,h) \;
S2. Search for writing skills online (5\,h) \;
S3. Take a writing class (10\,h) \\
S4. \textbf{Practice writing an hour a day (90\,days)} \;
S5. Write a book (365\,days) \;
S6. Shop the book around (90\,days) \;
S7. Get famous (365\,days)

\smallskip
\textbf{Stated constraints:} S1$\to$S3, \; S2$\to$S3, \; S3$\to$S4, \; S4$\to$S5, \; S5$\to$S6, \; S6$\to$S7

\smallskip
\textbf{Gold makespan:} 78,732,000\,s \quad \textbf{Predicted makespan:} 71,280,000\,s
\end{tcolorbox}

\noindent The model's output contains all six correct dependencies.
The sole error is step S4: ``90 days'' is parsed as $90 \times 3{,}600 = 324{,}000$\,s (90 hours) instead of $90 \times 86{,}400 = 7{,}776{,}000$\,s (90 days), which is an off-by-$24\times$ unit confusion.
Because S4 lies on the unique critical path S3$\to$S4$\to$S5$\to$S6$\to$S7, this single mistake cascades into a 7.5-million-second underestimate.

\paragraph{Example 4: Incorrect PDDL formalization introduces a circular dependency.}

The stated constraints for this 5-step task are S3$\to$S1, S2$\to$S3, S2$\to$S4, S1$\to$S5.
The constraint S3$\to$S1 means step 3 must finish \emph{before} step 1 starts, so \texttt{do\_step1} correctly lists \texttt{(s3\_done)} as a precondition.
GPT-5-mini's PDDL, however, adds \texttt{(s1\_done)} to \texttt{do\_step3}'s preconditions as well:

\begin{tcolorbox}[colback=gray!8, colframe=gray!50, title={Task: \textit{Make Farina} (GPT-5-mini, AsyncPlan-XXL $N{=}5$)}, fonttitle=\bfseries\small, left=4pt, right=4pt, top=4pt, bottom=4pt]
\begin{alltt}\scriptsize
; Measure dry farina grains (30 s)
(:durative-action do\_step1
  ; S3->S1 (correct)
  :condition (and ... (at start (s3\_done))))

(:durative-action do\_step3  ; Whisk grains (2 min)
  ; S2->S3 (correct)
  :condition (and ... (at start (s2\_done))
  ; S1->S3 is spurious
                      (at start (s1\_done))))
\end{alltt}
\end{tcolorbox}

\noindent This creates an unsatisfiable cycle: \texttt{do\_step1} requires \texttt{s3\_done} (step 3 must finish first), yet \texttt{do\_step3} requires \texttt{s1\_done} (step 1 must finish first).
OPTIC detects the mutual dependence, reports a syntax error, and returns no plan.
The gold makespan is 810\,s; the predicted value is \texttt{None}.

\paragraph{Example 5: Incomplete PDDL goal and missing dependencies.}\

\textit{Incomplete goal.}\;
PDDL2.1 Formalizer consistently omits intermediate completion predicates from the \texttt{:goal}.
For the Eggless Chocolate Cake task (Example~1), the problem file contains:

\begin{tcolorbox}[colback=gray!8, colframe=gray!50, title=Problem PDDL goal (Eggless Chocolate Cake), fonttitle=\bfseries\small, left=4pt, right=4pt, top=4pt, bottom=4pt]
{\scriptsize\ttfamily
(:goal (and (step\_done step1) (step\_done step2) (step\_done step3) (step\_done step4) (step\_done step5) (s5\_done)))
\textit{; only the terminal semantic predicate is present --- s1\_done through s4\_done are absent}
}
\end{tcolorbox}

\noindent The predicates \texttt{s1\_done} through \texttt{s4\_done} are absent from the goal.
Because the temporal planner only needs to satisfy \texttt{s5\_done} semantically, it is free to reach that predicate without ever producing the intermediate effects that the problem intended to enforce.

\textit{Missing dependencies.}\;
As graph size grows, Planner begins to drop entire edges from its dependency output.
Each omitted constraint allows the temporal planner to start a step earlier than the task permits, shortening the predicted makespan below the gold.
The supported-subset precedence-recall diagnostic is reported with their denominators in Table~\ref{tab:stepwise-diagnostics}.

\paragraph{Example 6: Planner schedules two items on the same station simultaneously (Robo Challenge eval error).}

In Robo Challenge, the Planner must respect physical station-capacity constraints: the fryer, grill, boiler, and marinator each have a fixed maximum concurrency.
Even when the overall plan structure is reasonable, the model often ignores whether a station is already in use.

\begin{tcolorbox}[colback=gray!8, colframe=gray!50, title={Task: \textit{Burger Station} (Gemini-3-Flash, Robo Challenge Hard)}, fonttitle=\bfseries\small, left=4pt, right=4pt, top=4pt, bottom=4pt]
\small
\textbf{Items requiring frying:} fish\,\#01 (raw$\to$fried, 8\,s), \; potato\,\#01 (cut$\to$fried, 8\,s)

\smallskip
\textbf{Station constraint:} fryer capacity\,$= 1$ (at most one fry action at a time)

\smallskip
\textbf{Submitted plan (partial):}
\begin{alltt}\scriptsize
0.000: (fry fish01)   [8.0]   <- fryer occupied until t=8
0.000: (cut potato01) [3.0]   <- cut finishes at t=3
; fryer still occupied by fish01!
3.000: (fry potato01) [8.0]
\end{alltt}

\textbf{Simulator error:} \texttt{t=3.000 START (fry potato01): station 'fryer' is full (capacity 1)}
\end{tcolorbox}

\noindent At $t{=}0$ the model correctly launches \texttt{fry fish01}.
After \texttt{cut potato01} finishes at $t{=}3$, it immediately launches \texttt{fry potato01}, without checking that the fryer is occupied until $t{=}8$.
The simulator rejects the plan at the first conflict.
A correct plan would delay \texttt{fry potato01} to $t{=}8$, achieving the 25\,s optimal makespan.

\paragraph{Example 7: CP-SAT Formalizer references a committed action in replanning (Online Robo Challenge spec error).}

Online Robo Challenge introduces dynamic events (new deliveries, priority changes) that trigger mid-execution replanning.
At each replanning stage, the model must output a new schedule for \emph{remaining} work only; any action already committed or in progress must not be re-encoded.
GPT-5-mini violates this constraint at Stage~2 of a 3-stage episode:

\begin{tcolorbox}[colback=gray!8, colframe=gray!50, title={Online Replanning Stage 2 ($t{=}8.17$\,s): \textit{Burger Station} (GPT-5-mini)}, fonttitle=\bfseries\small, left=4pt, right=4pt, top=4pt, bottom=4pt]
\small
\textbf{Context:} potato is \emph{currently boiling} (committed; finishes at $t{\approx}12$).
\texttt{mash\_potato} must not start until boiling is done.

\smallskip
\textbf{Model's Stage-2 scheduling JSON (excerpt):}
\begin{alltt}\scriptsize
"actions": [..., \{"id": "mash\_potato", ...\}, ...],
"dependencies": [
  \{"predecessor": "boil\_potato",  \textit{; <- not in actions!}
   "successor":   "mash\_potato"\},
  ; boil_potato is committed
  ...
]
\end{alltt}

\textbf{Spec error:} \texttt{dependency references unknown action: 'boil\_potato' -> 'mash\_potato'}
\end{tcolorbox}

\noindent The model correctly identifies that \texttt{mash\_potato} must wait for boiling to finish.
However, it encodes this as a dependency edge \texttt{boil\_potato}$\to$\texttt{mash\_potato}, referencing \texttt{boil\_potato} as if it were a new action, but \texttt{boil\_potato} is an ongoing committed action absent from the Stage-2 actions list.
The CP-SAT validator rejects the spec because dependency endpoints must be declared actions.
The correct encoding would use a release-time offset on \texttt{mash\_potato} derived from when the committed boil action is expected to finish.

\FloatBarrier
\section{Prompts}
\label{app:prompts}

This appendix reports the prompt templates used in our experiments. All three planning pipelines take the same input task but differ in their output: the Planner returns a plan directly, the PDDL2.1 Formalizer returns PDDL domain and problem files for OPTIC, and the CP-SAT Formalizer returns a compact scheduling JSON specification for CP-SAT. Figures~\ref{fig:planner_prompt}, \ref{fig:pddl_formalizer_prompt}, and~\ref{fig:cpsat_formalizer_prompt} show the respective prompts, Figures~\ref{fig:pddl_retry_prompt} and~\ref{fig:cpsat_retry_prompt} show the retry prompts for syntax and validation errors, and Figure~\ref{fig:nl_generation_prompt} gives the prompt used to rewrite abstract DAGs into natural language for AsyncPlan-XXL.

\subsection{AsyncHow and AsyncPlan-XXL Prompts}

AsyncHow and AsyncPlan-XXL instantiate the action-constrained setting: each
problem specifies step durations and ordering constraints, assumes infinite
parallel resources, and asks for the shortest possible makespan.

%--------------------------- Planner prompt ---------------------------%
\begin{figure*}[t]
\begin{tcolorbox}[colback=orange!8, colframe=orange!70!black,
    title=Prompt for Planner, fonttitle=\bfseries]
\small
\textbf{System prompt:} You are a helpful plan organizer.

\vspace{4pt}
\textbf{User prompt:}

\vspace{2pt}
\{question\}

\vspace{4pt}
Return ONLY valid JSON with this exact schema:

\vspace{2pt}
\texttt{\{}\\
\texttt{\ \ "actions": [}\\
\texttt{\ \ \ \ \{"id": "step1", "duration": 300\},}\\
\texttt{\ \ \ \ \{"id": "step2", "duration": 900\}}\\
\texttt{\ \ ],}\\
\texttt{\ \ "dependencies": [}\\
\texttt{\ \ \ \ \{"before": "step1", "after": "step2"\}}\\
\texttt{\ \ ],}\\
\texttt{\ \ "final\_answer\_seconds": 1200}\\
\texttt{\}}

\vspace{4pt}
\textit{Rules:}
\begin{enumerate}[leftmargin=*, nosep]
  \item Include one action for every listed step.
  \item Use ids exactly \texttt{"step1"}, \texttt{"step2"}, \ldots, matching the original step numbers.
  \item Convert every duration to seconds.
  \item Include every direct ordering constraint as a dependency.
  \item \texttt{"before"} is the step that must finish first; \texttt{"after"} is the step that starts later.
  \item \texttt{final\_answer\_seconds} is your computed shortest possible completion time under infinite resources.
  \item Do not include markdown, prose, or explanation.
\end{enumerate}
\end{tcolorbox}
\caption{\textbf{Prompt template for Planner on AsyncHow and AsyncPlan-XXL.}}
\label{fig:planner_prompt}
\end{figure*}

%--------------------------- PDDL Formalizer prompt ---------------------------%
\begin{figure*}[t]
\begin{tcolorbox}[colback=yellow!5!white, colframe=blue!55!black,
    title=Prompt for PDDL2.1 Formalizer, fonttitle=\bfseries]
\small
\textbf{System prompt:} You are a PDDL expert. Given an asynchronous planning
problem, write the domain and problem files in minimal PDDL2.1 format.

\vspace{3pt}
\textit{Encoding.} Use parameterless actions: no step type and no step objects.
Each action gets its own pair of predicates, \texttt{(X\_pending)} and
\texttt{(X\_done)}. \texttt{X\_done} is both the step-completed marker and the
semantic predicate used by successor actions.

\vspace{3pt}
\textit{OPTIC compatibility rules.}
\begin{enumerate}[leftmargin=*, nosep]
  \item Use only \texttt{:durative-actions} in \texttt{:requirements}; do not use \texttt{:typing}, \texttt{:negative-preconditions}, \texttt{:adl}, or \texttt{:disjunctive-preconditions}.
  \item Never use \texttt{(not ...)} in action preconditions. Use positive \texttt{\_pending} predicates in conditions, and delete them only in effects.
  \item Declare all predicates before use. Do not include \texttt{:types} or \texttt{:objects}; actions take no parameters.
  \item Use \texttt{(:durative-action ...)} with exactly one \texttt{:duration}, one \texttt{:condition}, and one \texttt{:effect}.
  \item Use \texttt{(at start ...)}, \texttt{(at end ...)}, or \texttt{(over all ...)} inside conditions and effects.
  \item Initialize all \texttt{\_pending} predicates in the problem \texttt{:init}.
\end{enumerate}

\vspace{3pt}
\textit{Semantic rules.}
\begin{enumerate}[leftmargin=*, nosep]
  \item Every action must consume its own \texttt{(X\_pending)} at start, produce its own \texttt{(X\_done)} at end, and require all direct predecessors' \texttt{\_done} predicates as \texttt{(at start ...)} conditions.
  \item The graph is a DAG, not a linear chain: if a step has multiple predecessors, enforce all of them as an AND-join.
  \item The problem goal must include all \texttt{\_done} predicates, one per action, so the temporal planner cannot skip steps.
  \item Every action duration must be exactly the duration stated in the problem, converted to seconds. Do not introduce one-second placeholder actions.
  \item The number of durative actions must exactly equal the number of listed steps.
\end{enumerate}

\vspace{3pt}
\textit{Dependency analysis.} For each step, ask what must be fully completed
before this step can start. Check explicit cues such as ``after'', ``once'',
``following'', and numbered constraints. A missing predecessor can let OPTIC
find an illegally short makespan.

\vspace{3pt}
\textit{Output format.} Return JSON with key \texttt{"responses"} containing a
list of dictionaries. Each dictionary includes \texttt{domain\_pddl},
\texttt{problem\_pddl}, and \texttt{step\_actions}, where
\texttt{step\_actions[i]} is the PDDL action name for Step \texttt{i+1}.

\vspace{4pt}
\textbf{User prompt:}

\vspace{2pt}
Here is an asynchronous planning problem.

\vspace{2pt}
\{question\}

\vspace{2pt}
The problem contains exactly \{n\_steps\} numbered steps. The corresponding step ids are:

\{step\_ids\}

\vspace{2pt}
Generate exactly one durative action for each listed step, and give the one-to-one mapping from these step ids to PDDL action names in \texttt{step\_actions}, ordered by step number.
\end{tcolorbox}
\caption{\textbf{Prompt template for the PDDL2.1 Formalizer on AsyncHow and AsyncPlan-XXL.}}
\label{fig:pddl_formalizer_prompt}
\end{figure*}

%--------------------------- CP-SAT Formalizer prompt ---------------------------%
\begin{figure*}[t]
\begin{tcolorbox}[colback=green!5!white, colframe=green!45!black,
    title=Prompt for CP-SAT Formalizer, fonttitle=\bfseries]
\small
\textbf{System prompt:} You are an expert scheduling formalizer.

\vspace{2pt}
Translate each natural-language async planning problem into a JSON scheduling
specification for a CP-SAT solver.

\vspace{3pt}
\textit{Rules:}
\begin{enumerate}[leftmargin=*, nosep]
  \item Each numbered step in the problem becomes exactly one action.
  \item Use \texttt{action="task"} for every action.
  \item Use the exact ids \texttt{step1}, \texttt{step2}, \ldots, \texttt{stepN} based on the step numbers in the problem text.
  \item Use item values identical to the action ids: \texttt{step1}, \texttt{step2}, \ldots, \texttt{stepN}.
  \item Duration must be the integer number of seconds for that step.
  \item Add a dependency only when the problem states that one step must happen before another. Do not add extra dependencies.
  \item Do not add stations, robots, or resources.
  \item Output only valid JSON with keys \texttt{actions} and \texttt{dependencies}.
  \item Before finishing, verify that every step appears exactly once, every dependency endpoint exists, durations are integer seconds, and there are no duplicate actions.
\end{enumerate}

\vspace{4pt}
\textbf{User prompt:}

\vspace{2pt}
Translate this planning problem into a CP-SAT scheduling JSON spec.

\vspace{2pt}
Problem:

\{question\}

\vspace{2pt}
The problem contains exactly \{n\_steps\} numbered steps. Your action ids must be:

\{step\_ids\}

\vspace{2pt}
Return only the JSON object.
\end{tcolorbox}
\caption{\textbf{Prompt template for the CP-SAT Formalizer on AsyncHow and AsyncPlan-XXL.}}
\label{fig:cpsat_formalizer_prompt}
\end{figure*}

%--------------------------- Retry prompts ---------------------------%
\begin{figure*}[t]
\begin{tcolorbox}[colback=yellow!5!white, colframe=blue!50!black,
    title=PDDL Formalizer Retry Prompt, fonttitle=\bfseries]
\small
The PDDL you generated caused the following error from the OPTIC planner:

\vspace{2pt}
\{error\}

\vspace{4pt}
Given the previous PDDL and the error message:
\begin{enumerate}[leftmargin=*, nosep]
  \item Analyze the error message and identify all issues; there may be more than one.
  \item Think step by step about how each issue relates to the PDDL syntax or semantics.
  \item Fix all issues and return the corrected domain and problem PDDL.
\end{enumerate}
\end{tcolorbox}
\caption{\textbf{Syntax retry prompt for the PDDL2.1 Formalizer.}}
\label{fig:pddl_retry_prompt}
\end{figure*}

\begin{figure*}[t]
\begin{tcolorbox}[colback=green!5!white, colframe=green!45!black,
    title=CP-SAT Formalizer Retry Prompt, fonttitle=\bfseries]
\small
Your previous scheduling JSON could not be parsed or validated.

\vspace{2pt}
Error:

\{error\}

\vspace{4pt}
Please return a corrected JSON object only. Remember:
\begin{itemize}[leftmargin=*, nosep]
  \item exactly one action per numbered step;
  \item action ids must be \texttt{step1..stepN};
  \item action must always be \texttt{"task"};
  \item durations must be integer seconds;
  \item dependencies may only reference existing action ids.
\end{itemize}
\end{tcolorbox}
\caption{\textbf{Validation retry prompt for the CP-SAT Formalizer.}}
\label{fig:cpsat_retry_prompt}
\end{figure*}

%--------------------------- Natural language generation prompt ---------------------------%
\begin{figure*}[t]
\begin{tcolorbox}[colback=gray!5, colframe=gray!60,
    title=Prompt for Natural-Language Rewriting (AsyncPlan-XXL), fonttitle=\bfseries]
\small
\textbf{System prompt:} You are a creative writer who specializes in turning
abstract asynchronous planning problems into concrete, realistic scenarios.

\vspace{2pt}
You will be given an abstract asynchronous planning problem that has a task
title, numbered steps with no names or durations, and ordering constraints
between steps. Your job is to fill in each step with a concrete activity name
and a realistic duration in natural language.

\vspace{3pt}
\textit{Rules:}
\begin{enumerate}[leftmargin=*, nosep]
  \item Keep the task title exactly as given.
  \item Keep the exact same number of steps.
  \item Copy all ordering constraints verbatim, with the same step numbers and wording.
  \item Copy the final question line verbatim.
  \item For each step, add a concrete activity-specific name and a reasonable duration in parentheses.
  \item The step names should form a coherent, realistic workflow for the task.
  \item Use diverse but plausible durations; not every step should use the same unit or take the same amount of time.
  \item The ordering constraints should make logical sense with the step names.
  \item Output only the rewritten problem. No commentary, markdown fences, or extra text.
\end{enumerate}

\vspace{4pt}
\textbf{User prompt:}

\vspace{2pt}
Here is an abstract asynchronous planning problem: \{abstract\_planning\_question\}
\end{tcolorbox}
\caption{\textbf{Prompt for the second stage of the AsyncPlan-XXL generation pipeline.} Abstract DAGs are rewritten into natural-language planning problems with concrete step names and realistic durations.}
\label{fig:nl_generation_prompt}
\end{figure*}

\subsection{Robo Challenge Prompts}

Robo Challenge uses the same three planning pipelines, but the input task now
contains executable cooking actions, item states, station capacities, optional
robot constraints, temporal lags, and optimization objectives. The prompts
therefore require the model to preserve exact item and robot identifiers and to
respect both temporal and executable-state constraints.

%--------------------------- Robo Planner prompt ---------------------------%
\begin{figure*}[t]
\begin{tcolorbox}[colback=orange!8, colframe=orange!70!black,
    title=Prompt for Planner on Robo Challenge, fonttitle=\bfseries]
\small
\textbf{System prompt:} You are an expert kitchen scheduler. Given a cooking
task, produce an optimal temporal plan that completes all required preparations
as fast as possible by running independent actions in parallel.

\vspace{3pt}
\textit{Output format.} Output one action per line:

\vspace{2pt}
\texttt{<time>: (<action> <item>) [<duration>]}

\vspace{2pt}
If the task lists robots, output:

\vspace{2pt}
\texttt{<time>: (<action> <robot> <item>) [<duration>]}

\vspace{3pt}
\textit{Rules:}
\begin{enumerate}[leftmargin=*, nosep]
  \item \texttt{<time>} is the start time in seconds; \texttt{<duration>} must match the task duration.
  \item Allowed actions are \texttt{grill}, \texttt{cut}, \texttt{fry}, \texttt{boil}, \texttt{toast}, \texttt{marinate}, \texttt{mash}, and \texttt{stack}.
  \item Use exact item names and, when present, exact robot names from the task.
  \item Independent actions may run in parallel, but an item cannot receive two actions at once.
  \item Respect action preconditions: for example, \texttt{marinate} requires \texttt{cut}, \texttt{mash} requires \texttt{boiled}, and task-specific cut-before-fry or marinate-before-grill requirements must be obeyed.
  \item Respect station capacities and robot exclusivity.
  \item Stack items only after all required processing is complete, and stack in the required bottom-to-top order.
  \item Enforce all temporal wait constraints and deadlines.
  \item For reward-maximization tasks, choose a feasible candidate subset under the deadline and inventory limits.
  \item Output only plan lines; no explanation or markdown.
\end{enumerate}

\vspace{4pt}
\textbf{User prompt:}

\vspace{2pt}
Produce a temporal plan for the following cooking task.

\vspace{2pt}
\{nl\}

\vspace{2pt}
IMPORTANT -- use these EXACT item names (with underscores) in your plan:

\{item\_names\}

\vspace{2pt}
If robots are listed in the task, include the robot argument in every action.
Output only the plan lines in the format shown.
\end{tcolorbox}
\caption{\textbf{Prompt template for Planner on Robo Challenge.}}
\label{fig:robo_planner_prompt}
\end{figure*}

%--------------------------- Robo PDDL prompt ---------------------------%
\begin{figure*}[t]
\begin{tcolorbox}[colback=yellow!5!white, colframe=blue!55!black,
    title=Prompt for PDDL2.1 Formalizer on Robo Challenge, fonttitle=\bfseries]
\small
\textbf{System prompt:} You are an expert AI planner. Translate a natural
language cooking task into PDDL2.1 with durative actions for the OPTIC temporal
planner.

\vspace{3pt}
\textit{Design pattern.}
\begin{enumerate}[leftmargin=*, nosep]
  \item Use a \texttt{(stackable <item>)} predicate to indicate that an item is ready to be stacked.
  \item Generate one separate durative action per item in the stack order, named \texttt{stack\_<item>}.
  \item Encode each station capacity with a \texttt{<station>\_free} predicate: an action requires it at start, deletes it at start, and restores it at end.
\end{enumerate}

\vspace{3pt}
\textit{Interface contract.}
\begin{enumerate}[leftmargin=*, nosep]
  \item Executable base action names must be only \texttt{grill}, \texttt{cut}, \texttt{fry}, \texttt{boil}, \texttt{toast}, \texttt{marinate}, \texttt{mash}, and \texttt{stack}.
  \item If an item-specific ground action is needed, name it \texttt{<base\_action>\_<exact\_item\_name>}; never invent names such as \texttt{cook}, \texttt{prepare}, \texttt{grill\_raw}, or \texttt{fry\_generic}.
  \item If robots are present, executable actions must include a robot argument and must require and release \texttt{(robot\_free ?r)}.
  \item Use exact item and robot names everywhere; never shorten names or insert underscores before numeric suffixes.
\end{enumerate}

\vspace{3pt}
\textit{PDDL rules.}
\begin{enumerate}[leftmargin=*, nosep]
  \item Use \texttt{(:requirements :durative-actions :typing)} and declare type \texttt{item}; also declare \texttt{robot} when robots are listed.
  \item Declare predicates for item states, \texttt{stackable}, \texttt{stacked}, \texttt{ready}, station-free predicates, and optional robot-free predicates.
  \item Processing actions consume their required input state and produce their output state; task-specific chains must update both conditions and effects, e.g., cut-before-fry uses \texttt{(cut ?i)}, not \texttt{(raw ?i)}.
  \item Stack actions require \texttt{(stackable <item>)} and, except for the first stack item, \texttt{(stacked <previous\_item>)}.
  \item Temporal waits are encoded with planner-only \texttt{wait\_*} durative actions plus completion and lag predicates.
  \item The goal includes all required processed states and all stacked items in the final stack order.
  \item Do not use negative preconditions, \texttt{or}, or \texttt{when}.
\end{enumerate}

\vspace{3pt}
\textit{Output format.} Return only valid JSON:

\vspace{2pt}
\texttt{\{"domain\_pddl": "<full domain PDDL>", "problem\_pddl": "<full problem PDDL>"\}}

\vspace{4pt}
\textbf{User prompt:}

\vspace{2pt}
Translate the following cooking task into PDDL2.1.

\vspace{2pt}
\{nl\}

\vspace{2pt}
IMPORTANT -- use these EXACT item names in all PDDL identifiers:

\{item\_names\}

\vspace{2pt}
If the task lists robots, use these EXACT robot names:

\{robot\_names\}

\vspace{2pt}
Output JSON with keys \texttt{"domain\_pddl"} and \texttt{"problem\_pddl"}.
\end{tcolorbox}
\caption{\textbf{Prompt template for the PDDL2.1 Formalizer on Robo Challenge.}}
\label{fig:robo_pddl_prompt}
\end{figure*}

%--------------------------- Robo CP-SAT prompt ---------------------------%
\begin{figure*}[t]
\begin{tcolorbox}[colback=green!5!white, colframe=green!45!black,
    title=Prompt for CP-SAT Formalizer on Robo Challenge, fonttitle=\bfseries]
\small
\textbf{System prompt:} You are an expert temporal scheduling formalizer.
Translate a natural-language cooking task into a structured scheduling JSON
model for a CP-SAT solver. The solver supports fixed-duration interval actions,
precedence dependencies, station capacities, and optional robot assignment; it
does not infer cooking semantics.

\vspace{3pt}
\textit{Rules:}
\begin{enumerate}[leftmargin=*, nosep]
  \item Allowed actions are \texttt{grill}, \texttt{cut}, \texttt{fry}, \texttt{boil}, \texttt{toast}, \texttt{marinate}, \texttt{mash}, and \texttt{stack}.
  \item Allowed non-stack stations are \texttt{grill}, \texttt{cutting\_board}, \texttt{fryer}, \texttt{boiler}, \texttt{toaster}, and \texttt{marinator}.
  \item Use exact item identifiers. Each action id must be \texttt{<action>\_<item>}; dependency endpoints must use existing action ids.
  \item Include all processing actions required to reach required states, and include stack actions only for items in the final stack order.
  \item Add dependencies for all multi-step chains, item-final-processing to stack, stack order, and explicit temporal waits.
  \item Do not add resource-order dependencies; CP-SAT handles station conflicts.
  \item If robots are listed, include \texttt{eligible\_robots} for every action.
  \item For candidate-reward tasks, output an optimization model rather than making every candidate action mandatory.
\end{enumerate}

\vspace{3pt}
\textit{Output format.} For ordinary scheduling tasks, return:

\vspace{2pt}
\texttt{\{"mode": "schedule", "actions": [...], "dependencies": [...]\}}

\vspace{2pt}
For reward optimization tasks, return:

\vspace{2pt}
\texttt{\{"mode": "optimize", "inventory\_limits": \{...\}, "candidate\_goals": [...]\}}

\vspace{4pt}
\textbf{User prompt:}

\vspace{2pt}
Translate this cooking task into the scheduling JSON model.

\vspace{2pt}
\{nl\}

\vspace{2pt}
Use these exact item identifiers in JSON:

\{item\_names\}

\vspace{2pt}
Return only the JSON object.
\end{tcolorbox}
\caption{\textbf{Prompt template for the CP-SAT Formalizer on Robo Challenge.}}
\label{fig:robo_cpsat_prompt}
\end{figure*}

\subsection{Online Robo Challenge Prompts}

Online Robo Challenge uses the same output formats as Robo Challenge, but
the prompt is conditioned on an execution snapshot. At each event time the model
receives the current time, completed actions, committed ongoing actions,
remaining work, per-item remaining chains, priority guidance, and deadline notes.

%--------------------------- Online context prompt ---------------------------%
\begin{figure*}[t]
\begin{tcolorbox}[colback=gray!5, colframe=gray!60,
    title=Shared Online Execution Context, fonttitle=\bfseries]
\small
\textbf{Context appended to the task description:}

\vspace{2pt}
\texttt{\#\# Online Execution Context}\\
Current time: \{now\} seconds.

\vspace{2pt}
Already completed before now:

\{completed\}

\vspace{2pt}
Committed ongoing actions:

\{ongoing\}

\vspace{2pt}
Remaining work only:

\{remaining\}

\vspace{2pt}
Per-item remaining action chains:

\{remaining\_actions\}

\vspace{2pt}
Rush-priority guidance:

\{priority\}

\vspace{2pt}
Deadline notes:

\{deadlines\}

\vspace{4pt}
\textit{Online instructions:}
\begin{itemize}[leftmargin=*, nosep]
  \item Output or encode only future work.
  \item Do not repeat completed actions or committed ongoing actions.
  \item Use full positive domain durations; never output zero-duration or partial remaining-time actions.
  \item Treat ongoing actions as occupying their resources until their committed finish times.
  \item If an item is already in the required prepared state, only plan later remaining actions such as stacking.
  \item Prioritize rush-critical deliveries before non-rush work whenever possible.
\end{itemize}
\end{tcolorbox}
\caption{\textbf{Shared online execution context appended to Online Robo Challenge prompts.}}
\label{fig:online_context_prompt}
\end{figure*}

%--------------------------- Online Planner prompt ---------------------------%
\begin{figure*}[t]
\begin{tcolorbox}[colback=orange!8, colframe=orange!70!black,
    title=Prompt for Planner on Online Robo Challenge, fonttitle=\bfseries]
\small
\textbf{System prompt:} You are an expert online kitchen scheduler. You will be
given the current state of a cooking episode after some actions may already
have started. Produce a valid future temporal plan for the remaining work only.

\vspace{3pt}
\textit{Strict formatting.} Output only raw plan lines. The first emitted
character must be a digit. If no robots are listed, use:

\vspace{2pt}
\texttt{<time>: (<action> <item>) [<duration>]}

\vspace{2pt}
If robots are listed, use:

\vspace{2pt}
\texttt{<time>: (<action> <robot> <item>) [<duration>]}

\vspace{3pt}
\textit{Rules:}
\begin{enumerate}[leftmargin=*, nosep]
  \item Times are relative to now; the first new action can start at \texttt{0.000}.
  \item Output only future actions that have not started yet.
  \item Do not repeat completed or ongoing actions; ongoing actions finish automatically.
  \item Respect remaining station occupancy, robot occupancy, action preconditions, deadlines, and inventory limits.
  \item Use exact item and robot names from the prompt.
  \item For optimization episodes, choose a feasible remaining subset under the deadline.
\end{enumerate}

\vspace{4pt}
\textbf{User prompt:}

\vspace{2pt}
Current time: \{now\} seconds

\vspace{2pt}
Plan the remaining work for this online cooking episode.

\vspace{2pt}
\{nl\}

\vspace{2pt}
Exact item names: \{item\_names\}

\vspace{2pt}
Already completed before now: \{completed\}

Committed ongoing actions: \{ongoing\}

Remaining work only: \{remaining\}

Per-item remaining action chains: \{remaining\_actions\}

Rush-priority guidance: \{priority\}

Deadline notes: \{deadlines\}

\vspace{2pt}
Output only the future suffix plan.
\end{tcolorbox}
\caption{\textbf{Prompt template for Planner on Online Robo Challenge.}}
\label{fig:online_planner_prompt}
\end{figure*}

%--------------------------- Online Formalizer prompts ---------------------------%
\begin{figure*}[t]
\begin{tcolorbox}[colback=blue!4!white, colframe=blue!55!black,
    title=Prompts for Online Formalizers, fonttitle=\bfseries]
\small
\textbf{Online PDDL Formalizer.} The system prompt is the Robo Challenge PDDL
Formalizer prompt in Figure~\ref{fig:robo_pddl_prompt}. The user prompt is the
same PDDL translation request, but \{nl\} includes the online execution context
from Figure~\ref{fig:online_context_prompt}. Additional online-specific
instructions require the model to generate PDDL only for remaining future work,
exclude committed actions, preserve full action durations, and encode ongoing
release times with planner-only \texttt{wait\_*} actions or equivalent lag
constraints.

\vspace{6pt}
\textbf{Online CP-SAT Formalizer.} The system prompt is the Robo Challenge
CP-SAT Formalizer prompt in Figure~\ref{fig:robo_cpsat_prompt}. The user prompt
is the same scheduling-JSON request, but \{nl\} includes the online execution
context. Additional online-specific instructions require the model to output
only remaining future work, avoid \texttt{*\_commit} or \texttt{*\_committed}
replacement actions, keep full positive durations, respect temporarily occupied
resources, and prioritize rush-critical deliveries.

\vspace{6pt}
\textbf{State-aware CP-SAT repair variant.} For the state-aware repair variant
used in Figure~\ref{fig:state-aware}, the LLM is not asked to regenerate a full
schedule specification. Instead it receives the current execution state and the
new event, and returns only an event delta:

\vspace{2pt}
\texttt{\{"delta": \{"add\_items": \{...\}, "add\_deliveries": [...],}\\
\texttt{\ \ "add\_temporal\_constraints": [...], "add\_candidate\_goals": [...],}\\
\texttt{\ \ "replace\_deliveries": null, "set\_deadline": null\}\}}

\vspace{2pt}
The corresponding user prompt contains current time, completed actions, ongoing
actions, and the JSON representation of the new online event, followed by:
\textit{Return the event delta JSON only.}
\end{tcolorbox}
\caption{\textbf{Prompt templates for the PDDL and CP-SAT Formalizers on Online Robo Challenge.}}
\label{fig:online_formalizer_prompts}
\end{figure*}

\end{document}